\documentclass[sigconf,screen]{acmart}

\AtBeginDocument{%
  \providecommand\BibTeX{{%
    \normalfont B\kern-0.5em{\scshape i\kern-0.25em b}\kern-0.8em\TeX}}}

\setcopyright{none}
\copyrightyear{2022}
\acmYear{2022}

\acmConference[ACM Multimedia]{ACM International Conference on Multimedia}{October 10--14,
  2022}{Lisbon, Portugal}

\usepackage[group-separator={,}]{siunitx}

\usepackage[capitalize]{cleveref}
\crefname{section}{Sec.}{Secs.}
\Crefname{section}{Section}{Sections}
\Crefname{table}{Table}{Tables}
\crefname{table}{Tab.}{Tabs.}
\Crefname{figure}{Figure}{Figures}
\crefname{figure}{Fig.}{Figs.}
\Crefname{equation}{Equation}{Equations}
\crefname{equation}{Eq.}{Eqs.}

\usepackage{xspace}
\makeatletter
\DeclareRobustCommand\onedot{\futurelet\@let@token\@onedot}
\def\@onedot{\ifx\@let@token.\else.\null\fi\xspace}

\def\eg{\emph{e.g}\onedot} 
\def\ie{\emph{i.e}\onedot} 
 
\def\etc{\emph{etc}\onedot} 
 
\def\etal{\emph{et al}\onedot}
\makeatother

\DeclareMathOperator*{\argmin}{arg\,min}

\usepackage[misc,geometry]{ifsym}
\usepackage{multirow}
\usepackage{makecell}
\usepackage{footmisc}
\usepackage{arydshln}
\usepackage{bm}
\usepackage{caption}
\usepackage[labelformat=simple]{subcaption}

\usepackage{enumitem}
\setlist[itemize]{noitemsep,nolistsep}

\renewcommand{\paragraph}[1]{\vspace{2pt} \noindent \textbf{#1}}
\newcommand{\red}[1]{\textcolor{red}{#1}}
\newcommand{\redb}[1]{\textcolor{red}{\textbf{#1}}}
\newcommand{\ublue}[1]{{\textcolor{blue}{#1}}}
\newcommand{\cirnum}[1]{{\large \textcircled{\small #1}}}
\newcommand{\RNum}[1]{\uppercase\expandafter{\romannumeral #1\relax}}

\newcommand{\hrp}{HRP\xspace}

\graphicspath{{supp_figs/}}

\begin{document}

\title{Real-World Blind Super-Resolution via Feature Matching with Implicit High-Resolution Priors}

\settopmatter{authorsperrow=4}
\author{Chaofeng Chen}
\authornote{Both authors contributed equally to this research.}
\email{chaofenghust@gmail.com}
\affiliation{%
  \institution{School of Informatics, Xiamen University}
  \city{Xiamen}
  \country{China}
}

\author{Xinyu Shi}
\authornotemark[1]
\email{x98shi@uwaterloo.ca}
\affiliation{%
  \institution{School of Informatics, Xiamen University}
  \city{Xiamen}
  \country{China}
}

\author{Yipeng Qin}
\email{QinY16@cardiff.ac.uk}
\affiliation{%
  \institution{School of Computer Science and Informatics, Cardiff Universiy}
  \city{Cardiff}
  \country{England}
}

\author{Xiaoming Li}
\email{csxmli@gmail.com}
\affiliation{%
  \institution{Faculty of Computing, Harbin Institute of Technology}
  \city{Harbin}
  \country{China}
}

\author{Tao Yang}
\email{yangtao9009@gmail.com}
\affiliation{%
  \institution{DAMO Academy, Alibaba Group}
  \city{Shenzhen}
  \country{China}
}

\author{Xiaoguang Han}
\email{hanxiaoguang@cuhk.edu.cn}
\affiliation{%
  \institution{SSE, The Chinese University of Hong Kong}
  \city{Shenzhen}
  \country{China}
}

\author{Shihui Guo}
\authornote{Corresponding author}
\email{guoshihui@xmu.edu.cn}
\affiliation{%
  \institution{School of Informatics, Xiamen University}
  \city{Xiamen}
  \country{China}
}
\renewcommand{\shortauthors}{Chen and Shi, et al.}

\begin{abstract}
A key challenge of real-world image super-resolution (SR) is to recover the missing details in low-resolution (LR) images with complex unknown degradations (\eg, downsampling, noise and compression). Most previous works restore such missing details in the image space. To cope with the high diversity of natural images, they either rely on the unstable GANs that are difficult to train and prone to artifacts, or resort to explicit references from high-resolution (HR) images that are usually unavailable. In this work, we propose Feature Matching SR (FeMaSR), which restores realistic HR images in a much more compact feature space. 
Unlike image-space methods, our FeMaSR restores HR images by matching distorted LR image {\it features} to their distortion-free HR counterparts in our pretrained HR priors, and decoding the matched features to obtain realistic HR images.
Specifically, our HR priors contain a discrete feature codebook and its associated decoder, which are pretrained on HR images with a Vector Quantized Generative Adversarial Network (VQGAN). Notably, we incorporate a novel semantic regularization in VQGAN to improve the quality of reconstructed images. 
For the feature matching, we first extract LR features with an LR encoder consisting of several Swin Transformer blocks and then follow a simple nearest neighbour strategy to match them with the pretrained codebook. In particular, we equip the LR encoder with residual shortcut connections to the decoder, which is critical to the optimization of feature matching loss and also helps to complement the possible feature matching errors.
Experimental results show that our approach produces more realistic HR images than previous methods.
Codes are released at \url{https://github.com/chaofengc/FeMaSR}.
\end{abstract}


\begin{teaserfigure}
  \includegraphics[width=\textwidth]{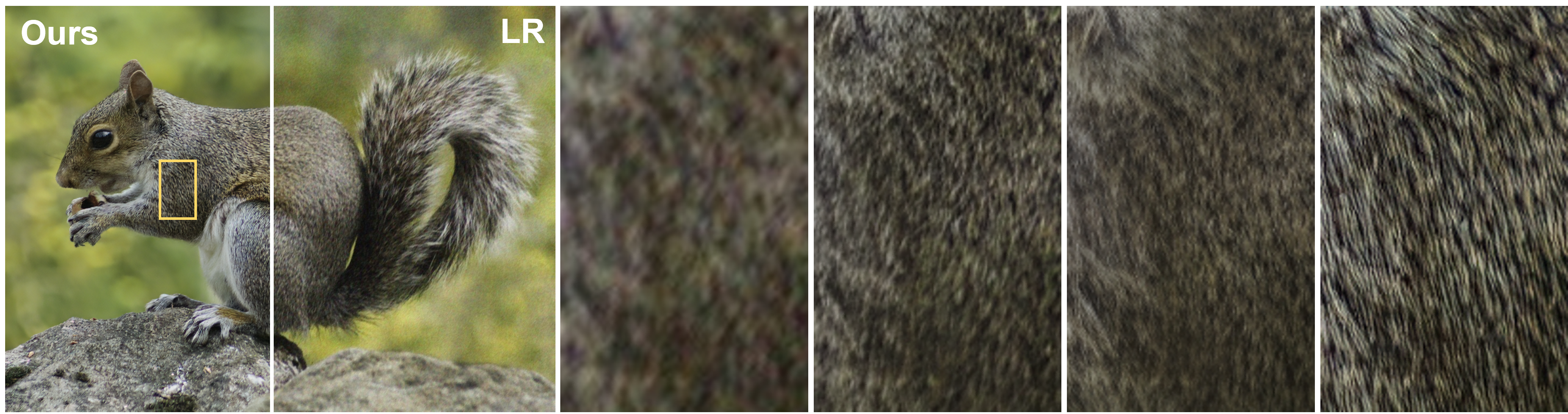}
  \\
  \makebox[0.35\linewidth]{\small Our SR Result / Low Resolution }
  \makebox[0.155\linewidth]{\small Bicubic}
  \makebox[0.155\linewidth]{\small Real-ESRGAN+ \cite{wang2021real}}
  \makebox[0.155\linewidth]{\small SwinIR-GAN \cite{liang2021swinir}}
  \makebox[0.155\linewidth]{\small FeMaSR (Ours)}
  \caption{Comparison between our FeMaSR and two latest works, Real-ESRGAN+ \cite{wang2021real} and SwinIR-GAN \cite{liang2021swinir} on a low resolution image with complex blind degradations. Our method can recover realistic hairs for the squirrel thanks to the implicit high-resolution priors. \red{Please zoom in for the best view.}}
  \label{fig:teaser}
\end{teaserfigure}

\maketitle

\section{Introduction} \label{sec:intro}

Single image super-resolution (SISR) is a fundamental task in low-level vision, aiming to restore high-resolution (HR) images from their low-resolution (LR) counterparts. 
Due to the incorporation of deep neural networks, previous works \cite{lim2017enhanced,zhang2018image,dai2019SAN,niu2020HAN,chen2020IPT,liang2021swinir} have made significant progress on {\it non-blind SR}, which assumes a known degradation process, \eg, bicubic downsampling.
However, these methods usually fail in real-world SR tasks where the degradations are unknown, \ie, \textit{blind SR}.

Blind SR is intrinsically an ill-posed problem because the complex and unknown distortions in the LR inputs have disrupted many details. 
Some works \cite{zhang2018learning,shocher2018zssr,gu2019ikc,zhou2019kmsr} exploited assumptions of the classical degradation model to explicitly estimate the blur kernel and noise.
As a result, most of them can only handle several simplified cases of the classical degradation model, and are a far cry from real-world SR solutions.
Other works \cite{zhang2019ranksrgan,wei2021unsupervised,zhang2021designing,wang2021real} resort to the synthesis power of Generative Adversarial Networks (GANs) to generate the missing textures.
Although effective, these approaches are prone to artifacts due to the notorious unstable GAN training.
Instead of ``guessing'' the missing textures, another line of research \cite{zheng2018crossnet,zhang2019image,yang2020ttsr,jiang2021c2match} takes advantages of reference images. Their performance is therefore determined by the reference HR images, which are not always available.
Addressing this issue, recent works \cite{pan2020dgp,wang2021towards} turned to implicit high-resolution priors implemented by {\it pretrained} GANs. Although bypassing the needs of explicit HR references, these methods are limited to the domain of the pretrained GANs (\eg, face images~\cite{chan2021glean,yang2021gan}) and cannot generalize to natural images with diverse contents\footnote{To our knowledge, all state-of-the-art GANs that can synthesize high-quality and high-resolution images are dedicated to a specific domain (\eg, StyleGAN~\cite{karras2020analyzing}).}.  

In this paper, we propose a novel SR framework based on feature matching, namely FeMaSR, for blind SR of real-world images. 
The distinct advantage of our framework is that it addresses the aforementioned limitations of previous works by matching LR features to a set of HR features in the pretrained implicit HR priors (\hrp).
Inspired by the recent VQ-VAE \cite{vqvae,vqvae2} and VQGAN \cite{esser2021taming}, we define our \hrp as the combination of a discrete codebook consisting of a pre-defined number of feature vectors and the corresponding pretrained decoder. The feature vectors contain the information of realistic textures that can be decoded into the target HR images.
In this way, we break blind SR into two sub-tasks: i) learning a high-quality HRP; ii) mapping the features of LR inputs to the codebook in \hrp for distortion removal and detail recovery. 
For the first sub-task, we pre-train our \hrp with a VQGAN that aims to reconstruct the input HR patches. However, instead of using the vanilla VQGAN, we incorporate semantic information into \hrp via L2 regularization with perceptual features from VGG19, thereby enhancing the correlation between semantics and codebook features. For the second sub-task, we follow SwinIR \cite{liang2021swinir} and utilize several swin transformer blocks to encode the LR inputs. The LR encoder is then trained with losses between LR features and ground truth HR features selected from the pretrained codebook. 
Especially, we found that the feature matching loss is difficult to optimize with fixed \hrp. To solve this problem, we introduce multi-scale residual shortcut connections from LR feature space to decoder features. These residual connections enable direct gradient flow from pretrained decoder to the LR encoder, thus making it easier to optimize the LR encoder. Besides, it also helps to complement the possible feature matching errors. Since \hrp contains rich semantic-aware HR information of natural images, the proposed FeMaSR is able to recover higher quality textures, see \cref{fig:teaser}. Our contributions can be summarized as follows:

\begin{itemize}
    \item We propose a novel framework FeMaSR for blind SR using \hrp encoded by a pretrained VQGAN network. Compared with previous works, the FeMaSR formulates SR as a feature matching problem between LR features and distortion-free HR priors, and therefore enables the generation of more realistic images with less artifacts for real-world SR.
    \item We introduce semantic regularization for the pretrain of semantic-aware \hrp. Such a regularization enhances the correlation between semantics and \hrp, thereby facilitating the generation of more realistic textures.
    \item We design a LR encoder with residual shortcut connections to the \hrp for feature matching. The proposed framework can better match the LR features with distortion free HR features, and also complement the matching errors.  
\end{itemize}

\begin{figure*}[t]
   \includegraphics[width=1.\linewidth]{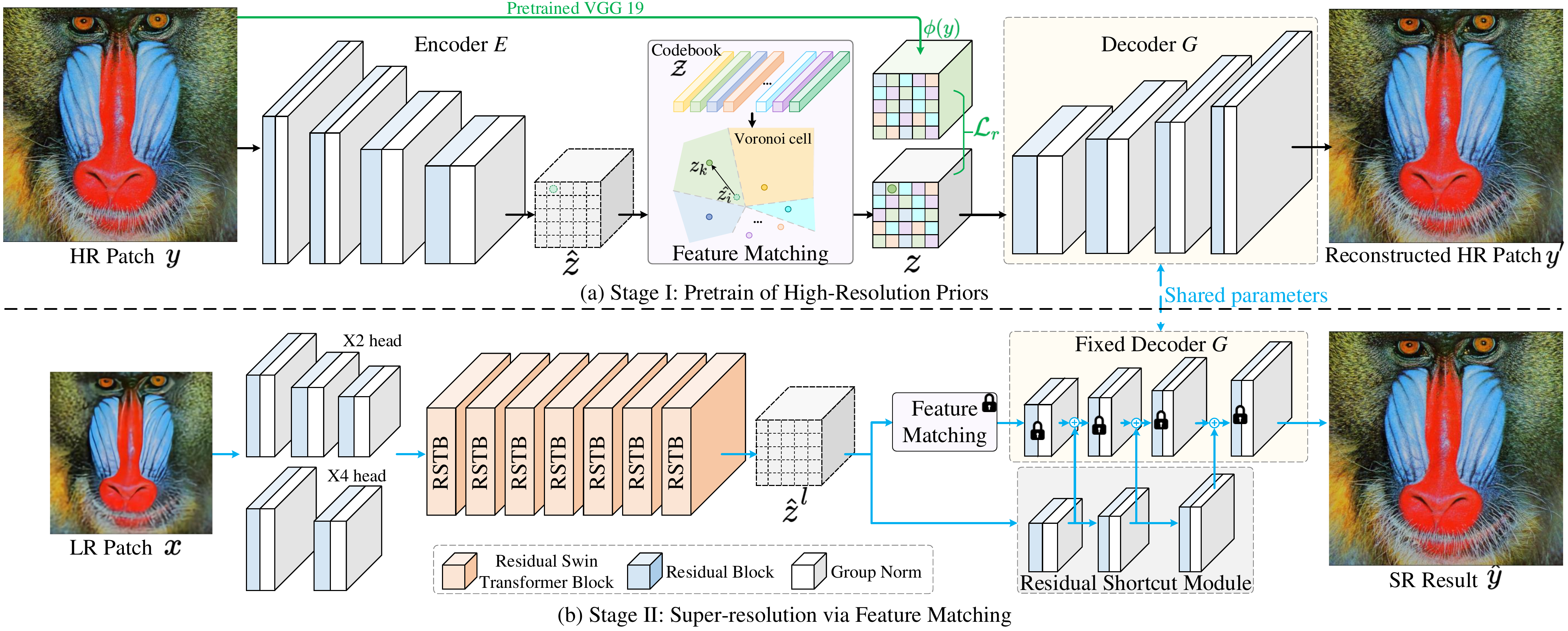}
   \caption{Framework of the proposed FeMaSR. It contains two stages: pretrain of high-resolution prior, and super-resolution via feature matching. We first pretrain a VQGAN to learn an implicit representation of high-resolution patches, \ie, the codebook $\mathcal{Z}$ and decoder $G$. Then the LR encoder $E_l$ is optimized to find the best matching features of the LR inputs $\bm{x}$ in the codebook $\mathcal{Z}$. Since $\mathcal{Z}$ and $G$ are pretrained to reconstruct high resolution patches, FeMaSR is able to generate clearer results with less artifacts.}
\label{fig:framework}
\end{figure*}

\section{Related Work}
\paragraph{Single Image Super-Resolution (SISR)} Starting from the pioneer SRCNN~\cite{dong2014learning}, deep neural networks have dominated the design of modern SR algorithms. Since then, various network architectures have been proposed to improve the performance of SISR. For example, Kim \etal \cite{kim2016accurate} proposed a deep version of SRCNN, named VDSR. Thanks to the residual \cite{he2016deep} and residual dense blocks \cite{huang2017densely} that enable training deeper and wider networks, EDSR \cite{lim2017enhanced} and RDN \cite{zhang2018residual} were proposed and boosted the performance of SISR. After that, the attention mechanism is also introduced to SISR, such as channel attention \cite{zhang2018image}, spatial attention \cite{niu2020HAN,ChenSPARNet}, non-local attention \cite{zhang2019rnan}, \etc. Latest works \cite{chen2020IPT,liang2021swinir} achieve state-of-the-art performance by employing vision image transformers \cite{liu2021swin}. These models are trained and evaluated in a non-blind manner, \eg, bicubic downsampling and blurring with known parameters, thereby making it difficult to generalize to SISR with the same degradation type but unseen parameters, let alone those with other degradation types.
Addressing this issue, Zhang \etal developed a series of methods \cite{zhang2018learning,zhang2019deep,zhang2020deep} for conditional image restoration, where users can control the outputs by changing the conditioned degradation parameters. 

\paragraph{Blind SISR} Upon the performance saturation of non-blind SISR, recent works turned to the more challenging real-world SISR with unknown degradation ({\it a.k.a.} blind SISR). In general, they model complex real-world degradations in either an implicit or an explicit way. 
Between them, implicit methods \cite{fritsche2019frequency,wei2021unsupervised,wan2020bringing,maeda2020unpaired,wang2021unsupervised} aim to learn a degradation network from real-world LR images. In the absence of corresponding ground truth HR images, most of them employed unsupervised image-to-image translation (\eg, Cycle-GAN \cite{zhu2017unpaired}) while some recent works \cite{zhang2021blind} resort to contrastive learning. On the contrary, explicit methods aim to synthesize ``real'' LR images by a manually designed degradation process.
Specifically, BSRGAN\cite{zhang2021designing} and Real-ESRGAN\cite{wang2021real} describe different ways to improve the common image degradation pipeline. Both of them demonstrate much better visual quality than implicit methods in blind SISR. Nevertheless, both implicit and explicit methods rely on the generative power of GANs to generate textures. However, GANs are known to have difficulties in distinguishing some real-world textures from similar degradation patterns, which usually lead to unrealistic textures or over-smoothed regions in the resulting HR images. 

\paragraph{Prior-based SISR} 
Since SISR is intrinsically an ill-posed problem, prior-based SISR methods take advantages of extra image priors either explicitly or implicitly. Methods based on explicit prior ({\it a.k.a.} RefSR) rely on one or multiple reference HR images which share the same or similar content with the input LR image. To locate the best reference images, various approaches were proposed, including cross-scale correspondences \cite{zheng2018crossnet}, texture transfer \cite{zhang2019image}, transformer network \cite{yang2020ttsr}, teacher-student \cite{jiang2021c2match}, internal graph \cite{zhou2020cross}, \etc. 
Li \etal \cite{li2018learning,li2020blind,li2020enhanced} narrow the image space to faces and achieve impressive performance. Although effective, explicit priors (\ie, HR reference images) are not always available for a given real-world LR image. Therefore, prior-based SISR is more promisingly achieved with a prior distribution (\ie, implicit prior) learnt from a large amount of HR images through GANs or VAEs. Menon \etal \cite{menon2020pulse} first proposed to upscale LR faces by searching the latent space of a pretrained StyleGAN generator \cite{karras2020analyzing}. 
Gu \etal \cite{gu2020image} improved it by introducing more latent codes. Pan \etal \cite{pan2020exploiting} exploited a BigGAN generator \cite{brock2018biggan} as a prior for versatile image restoration. Although these methods can generate realistic images, they all contain a time-consuming optimization process. Addressing this issue, \cite{yang2021gan,wang2021towards,chan2021glean} propose to learn a posterior distribution with a pretrained StyleGAN generator. Specifically, they learn an encoder to project LR images to a latent space shared with the pretrained generator that outputs HR images. Although this approach demonstrates exciting performance for face SR, it hardly works for natural images because learning a GAN for natural images remains a challenging task. In this work, we address the above-mentioned challenge following VQGAN \cite{esser2021taming} that shows outstanding performance in natural image synthesis and can be regarded as high-quality priors for image synthesis.

\section{Methodology} \label{sec:method}
\subsection{Framework Overview} \label{sec:framework}

Given an input LR image $\bm{x}$ with unknown degradations, we aim to restore the corresponding high-resolution image with realistic textures. As shown in \cref{fig:framework}, we employ a two-stage framework to pretrain the High-Resolution Priors (\hrp) and conduct feature matching sequentially:
\begin{itemize}
    \item {\bf Stage \RNum{1}, Pretraining of High-Resolution Priors.} 
    We use HR patches to pretrain a VQGAN~\cite{esser2021taming} consisting of an encoder $E$, a discrete codebook $\mathcal{Z}$, and a decoder $G$. Inspired by \cite{wang2018sftgan}, we train the VQGAN with semantic guidance that enhances the correlation of textures and semantics. We call the codebook $\mathcal{Z}$ and decoder $G$ \hrp. After pretraining, our \hrp approximately encodes the complete information of HR images and allows the reconstruction of them by feeding their corresponding feature codes $z \in \mathcal{Z}$ to $G$.
    \item {\bf Stage \RNum{2}, Super-Resolution via Feature Matching.} 
    Given the \hrp (\ie, $\mathcal{Z}$ and $G$) obtained in Stage \RNum{1}, we argue that blind SR is equivalent to a feature matching problem that aims to match the feature codes of LR inputs $\hat{z}^l$ to those of their HR counterparts $z \in \mathcal{Z}$. By feeding $G$ with the correctly matched HR feature codes $z$, we can obtain the clean and realistic HR images required in blind SR.
    To address the optimization challenges posed by the quantization process of VQGAN, we further propose the incorporation of a residual shortcut module to the LR encoder. This not only facilitates training but also complements the feature matching errors, which further boosts the quality of the resulting HR images.
\end{itemize}
Details are described in the following sections.

\subsection{Pretraining of High-Resolution Priors} \label{sec:hrp}

We first make a brief review of VQGAN. As illustrated in \cref{fig:framework}, the input HR image $\bm{y} \in \mathbb{R}^{H\times W \times 3}$ is first passed through the encoder $E$ to produce its output feature $\hat{z} = E(\bm{y}) \in \mathbb{R}^{h\times w \times n_z}$, where $n_z$ is the feature dimension. Then the discrete representation of $\hat{z}$ is calculated by finding the nearest neighbours of each element $\hat{z}_i \in \mathbb{R}^{n_z}$, in the codebook $\mathcal{Z} \in \mathbb{R}^{K \times n_z}$ as follows:
\begin{equation}
    z_i = \mathcal{Z}_k, \quad k = \argmin_j \| \hat{z}_i - \mathcal{Z}_j \|_2, \label{eq:quantize}
\end{equation}
where $z \in \mathbb{R}^{h \times w \times n_z}$, $K$ is the codebook size, $i\in \{1,2,\ldots,h\times w\}$, and $j \in \{1,2,\ldots,K\}$. After that, $\bm{y}$ is reconstructed by $z$ with the decoder $G$:
\begin{equation}
    \bm{y}' = G(z) \approx \bm{y},
\end{equation}
Since the feature quantization operation of \cref{eq:quantize} is non-differentiable, we follow \cite{vqvae,esser2021taming} and simply copy the gradients from $G$ to $E$ for backpropagation. Therefore, the model and codebook can be trained end-to-end with the following objective function:
\begin{align}
    \mathcal{L}_{VQ}(E, G, \mathcal{Z}) = \| \bm{y}' - \bm{y}\|_1 &+ \|\text{sg}[\hat{z}] - z \|_2^2 \nonumber \\ 
    & + \beta \| \text{sg}[z] - \hat{z} \|_2^2, \label{eq:vqgan}
\end{align}
where $\text{sg}[\cdot]$ is the stop-gradient operation, and $\beta=0.25$ according to \cite{vqvae,esser2021taming}. With the pretrained VQGAN, any high resolution images $\bm{y}$ from the training set can be reconstructed with their corresponding feature vectors in $\mathcal{Z}$ and the decoder $G$.
We therefore call them \hrp in this work.

\paragraph{Semantic Guidance} 
As indicated by the vanilla setting in \cref{eq:vqgan}, the codebook $\mathcal{Z}$ is learned purely by gradient descent where similar patterns are clustered independent of their semantics. Meanwhile, Wang \etal \cite{wang2018sftgan} pointed out that semantic guidance leads to better texture restoration. This motivates us to incorporate semantic information in the pretraining of VQGAN. 
To be specific, we regularize the training of codebook $\mathcal{Z}$ with perceptual features from a pretrained VGG19 network by adding a regularization term $\mathcal{L}_{r}$ to $\mathcal{L}_{VQ}$ and have
\begin{equation}
    \mathcal{L}'_{VQ} = \mathcal{L}_{VQ} + \mathcal{L}_{r} = \mathcal{L}_{VQ} + \gamma \|\mathrm{CONV}(z) - \phi(\bm{y}) \|_2^2
\end{equation} 
where $\mathrm{CONV}$ denotes a simple convolution layer to match the dimension of $z$ and $\phi{(\bm{y})}$, $\phi$ denotes the pretrained VGG19, and $\gamma$ is a weighting factor empirically set to $0.1$. Note that we follow \cite{esser2021taming} and also use perceptual loss and adversarial loss in the pretraining. 

In summary, our semantic-guided \hrp pretraining encourages the texture restoration to be conditioned on semantics, thereby enabling the restoration of more realistic and natural textures.

\subsection{Super-Resolution via Feature Matching}

With the pretrained \hrp, \ie, $\mathcal{Z}$ and $G$, the SR task is turned into a feature matching problem between LR inputs $\bm{x}$ and $\mathcal{Z}$. Denote the LR encoder as $E_l$, the problem can be formulated as
\begin{equation}
\argmin_{\theta} \mathcal{L}(G(\mathbf{q}[E_l(\bm{x}, \theta), \mathcal{Z}]), \bm{y}), \label{eq:feature_match}
\end{equation}
where $\theta$ is the learnable parameter of $E_l$, $\mathbf{q}[\cdot]$ denotes the feature matching process same as \cref{eq:quantize}, and $\mathcal{L}$ denotes the loss functions (which will be described in the following section). We first make a brief discussion about why we want to transform the SR task to a feature matching process and how can it help: 

As we know, image degradation is inherently a one-to-many mapping subject to different types and levels of degradation.
From a mathematical point of view, these degradations can be regarded as offsets of high-quality local features in some feature space, where the type and level of degradation correspond to the direction and distance of the offset respectively.
Such offsets overlap with each other, thereby making it difficult to find the correct high-quality correspondence of a degraded feature in the feature space. 
Heuristically, we address this challenge by mapping a degraded feature to its Euclidean nearest neighbhour in a given set of pre-defined high-quality features (\ie, the pretrained codebook $\mathcal{Z}$).
Intuitively, the codebook with discrete features partitions the feature space into non-overlapping cells that form a {\it degradation-based Voronoi diagram}.
As demonstrated in \cref{fig:framework}, we define the $K$ feature vectors $z_k$ in $\mathcal{Z}$ as the centers of $K$ Voronoi cells. 
Given an LR feature $\hat{z}^l_i$, we compute the Euclidean distance between $\hat{z}_i^l$ and all centers $z_k$ to determine which cell  $\hat{z}^l_i$ belongs to\footnote{In some rare cases (of zero probability),  $\hat{z}^l_i$ has the same nearest distances to multiple $z_k$, \ie, on the boundary of the Voronoi cells. In these cases, we randomly map  $\hat{z}^l_i$ to one of the centers.}, \ie, which $z_k$ it maps to. In this way, realistic and rich textures can be generated as the decoder inputs are mapped to expressive HR features $z_k$ instead of the raw LR features  $\hat{z}^l_i$.

Despite the advantages of feature matching, the optimization of \cref{eq:feature_match} is quite challenging because of the complex LR inputs. For this purpose, we introduce a powerful LR encoder $E_l$ consisting of two parts: feature extraction module and residual shortcut module. 

\paragraph{Feature Extraction} As shown in \cref{fig:framework}, the design of the feature extraction module basically follows SwinIR \cite{liang2021swinir}. It is composed of a shallow feature extraction head and a deep feature extraction block. The deep feature extraction block applies the same stack of residual swin transformer layers as SwinIR, while the shallow feature extraction block is slightly different. Since the \hrp is fixed, the final upscaling factor $S_{up}$ of the input LR image is controlled by the downscaling factor $S_{down}$ of the shallow feature encoder block. 
In this work, we have $S_{up} = S_{down} \times 8$ as the decoder $G$ upscales $z \in \mathbb{R}^{h\times w}$ by $\times8$.
Denote the feature extraction module as $H_F$, we have:
\begin{equation}
    \hat{z}^l = H_F(\bm{x}),
\end{equation}
where $\hat{z}^l \in \mathbb{R}^{h\times w\times n_z}$ are the LR features used for feature matching.

\paragraph{Residual Shortcut Module} To better utilize the \hrp, we further introduce multi-scale residual connections between $\hat{z}^l$ and the decoder $G$, as shown in \cref{fig:framework}. To be specific, we use several upsampling blocks $H_{up}$ to upscale LR features $\hat{z}^l$ and add them as residuals to the decoder $G$, \ie,
\begin{align}
    & f_0 = z, \hat{f_0} = \hat{z}^l \\
    & f_i = G_{up}^i(f_{i-1}) + H_{up}^i(\hat{f}_{i-1}), i \in \{1, 2, 3, ...\}
\end{align}
where $G_{up}^i$ and $H_{up}^i$ are the $i$-th upsampling blocks in $G$ and $H_F$ respectively, $f_{i-1}$ and $\hat{f}_{i-1}$ are the input features to them.

Our residual shortcut module has two main benefits.
First, it sidesteps the non-differentiable quantization process in VQGAN, thus allowing gradients to be backpropagated directly from $G$ to $E_l$, which greatly eases the optimization difficulty.
Second, we observed that these extra residual connections have also learned to complement the potential errors in feature matching and can further boost the performance of blind SR.

\subsection{Training Objectives} \label{sec:losses}

The gradients to update $E_l$ come from three parts: feature matching losses, image reconstruction losses, and adversarial loss. 

\paragraph{Feature Matching Loss} 
This loss is dedicated to the training of $E_l$.
We first obtain the ground truth latent representation of $\bm{y}$, \ie, $z_{gt}=\mathbf{q}[E(\bm{y}), \mathcal{Z}]$, and then calculate the L2 loss and the Gram matrix loss for LR features
\begin{equation}
    \mathcal{L}_{fema} = \beta \| \hat{z}^l - z_{gt}\|_2^2 + \alpha \| \psi({\hat{z}^l} - \psi(z_{gt})) \|_2^2,
\end{equation}  
where $\psi$ calculates the Gram matrix of features, and $\alpha$ is its weight. The Gram matrix loss, also called style loss, has been shown to be helpful to restore textures \cite{gondal2018unreasonable}.

\paragraph{Reconstruction Loss} We follow \cite{wang2021towards,esser2021taming} and employ L1 and perceptual losses as our reconstruction loss, formulated as
\begin{equation}
    \mathcal{L}_{rec} = \lambda_{L1} \| \hat{\bm{y}} - \bm{y} \|_1 +  \lambda_{per} \| \phi(\hat{\bm{y}}) - \phi(\bm{y}) \|_2^2
\end{equation}
where $\phi$ is a pretrained VGG-16 network, $\lambda_{L1}$ and $\lambda_{per}$ are weights of the L1 and perceptual losses respectively.

\paragraph{Adversarial Loss} Although our \hrp already contains rich texture information, we still need an adversarial loss to help us find better-matching features in the feature matching process. We follow \cite{wang2021real} and adopt a U-Net discriminator $D$ with spectral normalization \cite{miyato2018spectral}. Similar to \cite{ChenPSFRGAN}, we use a hinge loss and define the generator loss as
\begin{equation}
    \mathcal{L}_{adv} = \lambda_{adv} \sum_i - \mathbb{E}[D(\hat{\bm{y}_i})]
\end{equation} 
For simplicity, the discriminator loss is omitted here. 

\paragraph{Overall Loss} The overall loss is defined as
\begin{equation}
    \mathcal{L}_{total} = \mathcal{L}_{fema} + \mathcal{L}_{rec} + \mathcal{L}_{adv}
\end{equation}
where the weights for each loss are set as: $\alpha = \lambda_{L1} = \lambda_{per} = 1, \beta = 0.25, \lambda_{adv}=0.1$.

\section{Implementation Details}
\subsection{Datasets and Evaluation Metrics}

\paragraph{Training Dataset} We follow BSRGAN \cite{zhang2021designing} and build a training set that includes DIV2K \cite{DIV2K}, Flickr2K \cite{lim2017enhanced}, DIV8K \cite{gu2019div8k} and \num{10000} face images from FFHQ \cite{karras2017progressive}. We use the following ways to generate the training patches: (1) crop non-overlapping $512\times512$ patches; (2) filter patches with few textures; (3) for well-aligned faces in FFHQ, we perform random resize with scale factors between $[0.5, 1.0]$ before cropping to avoid content bias. More details are provided in the supplementary material. The final training dataset contains \num{136205} HR patches of size $512\times512$. We use the same degradation model as BSRGAN\footnote{\label{fn_bsrgan}\url{https://github.com/cszn/BSRGAN}} to generate corresponding LR images.  

\paragraph{Synthetic Testing Dataset} To ensure a fair comparison, we use a mixed degradation model of two recent works BSRGAN and Real-ESRGAN, denoted as \texttt{bsrgan\_plus}\footref{fn_bsrgan}, to generate LR testsets for DIV2K validation set and 5 classical benchmarks, \ie, Set5, Set14, BSD100, Urban100 and Manga109. The diversity of test images guarantees a comprehensive evaluation of model performance.

\paragraph{Real-world Testing Dataset} We test our model on three recent real-world benchmarks, including RealSR \cite{wang2021towards}, DRealSR \cite{wei2020cdc} and DPED-iphone \cite{ignatov2017dslr}. We test models with an upscale factor of 4 for these real-world datasets. Images from RealSR and DRealSR are captured by DSLR cameras, and contain \num{100} and \num{93} images respectively. DPED-iphone includes \num{100} LR images captured by smartphone cameras. The LR images in DPED-iphone are usually more corrupted than those from RealSR and DRealSR.

\paragraph{Evaluation Metrics} For synthetic test datasets with ground truth images, we employ the well-known perceptual metric, LPIPS \cite{zhang2018perceptual} score, to evaluate the perceptual quality of generated images. We also report the results of the widely used PSNR, SSIM scores for references. For real-world benchmarks, there are usually no ground truth images, therefore we adopt the well-known no reference metric NIQE score for quantitative comparison. 

\subsection{Training Details}

In both \hrp pretraining and SR training, we use the Adam optimizer \cite{kingma2014adam} with $\beta_1=0.9$ and $\beta_2=0.99$. The learning rates for the generator and discriminator are set to $0.0001$ and $0.0004$, respectively, throughout training. During the feature-matching stage, the codebook $\mathcal{Z}$ and decoder $G$ are fixed. Both \hrp and SR networks are trained with a batch size of 16, and the HR image size is fixed at $256\times256$ for both $\times2$ and $\times4$ upscaling factors. We implement our model in PyTorch \cite{pytorch}. The \hrp pretraining stage takes about 3 days on two GeForce RTX 3090 GPUs, and the SR stage takes about 4 days on the same hardware.

\section{Experiments}

\subsection{Visualization of \hrp}

\begin{figure}[t]
     \centering
     \begin{subfigure}[t]{0.49\linewidth}
         \centering
         \includegraphics[width=0.98\linewidth]{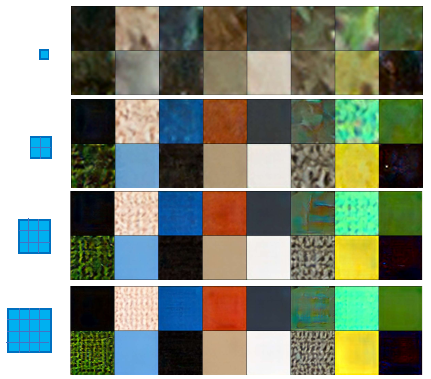}
         \caption{Textures generated with tiled single code. The tiled feature size are: $1\times1$, $2\times2$, $3\times3$, $4\times4$ (from top to bottom)}
         \label{fig:code_vis_single}
     \end{subfigure}
     \hfill
     \begin{subfigure}[t]{0.49\linewidth}
         \centering
         \includegraphics[width=0.98\linewidth]{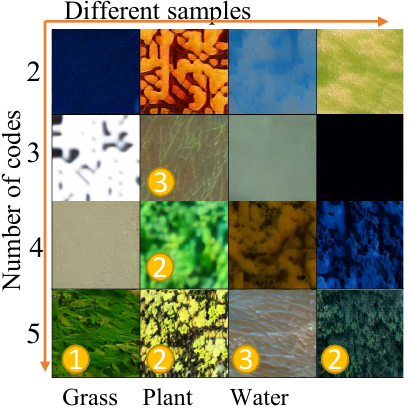}
         \caption{Textures generated with random combination with different number of codes. The size of combined feature map is $16\times16$.}
         \label{fig:code_vis_multi}
     \end{subfigure}
     \caption{Visualization of texture priors encoded with pretrained codebook $\mathcal{Z}$. Semantic textures emerge when different codes are combined , such as \cirnum{1} grass, \cirnum{2} plant and \cirnum{3} water.} \label{fig:code_vis}
\end{figure}

\begin{figure*}[p]
    \begin{minipage}{1.\linewidth}
    \centering
    \includegraphics[width=1.0\linewidth]{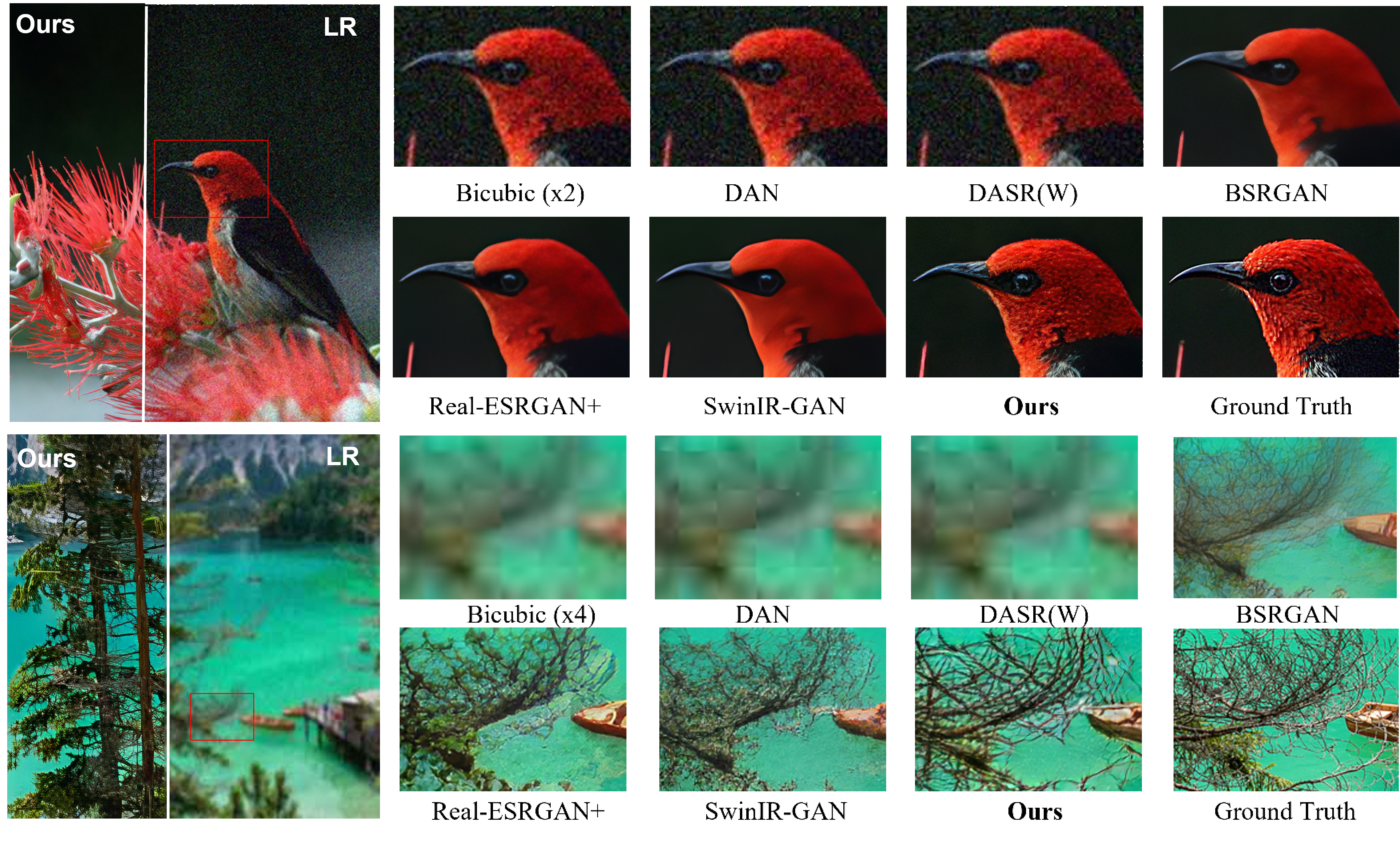}
    \caption{Visual comparisons on two examples from synthesize benchmarks with upscale factor of 2 (first row) and 4 (second row).Thanks to the HRP, our model is able to restore realistic and faithful textures even when the inputs are severely corrupted. As for the competitive works, some have difficulties to remove degradation, \ie, DAN and DASR(W), and the others generate artifacts or tend to be oversmooth, \ie, BSRGAN, Real-ESRGAN+, SwinIRGAN. \red{Please zoom in for best view.}} \label{fig:synthetic_results}
    \end{minipage}
    \begin{minipage}{1.\linewidth}
        \centering
        \includegraphics[width=1.0\linewidth]{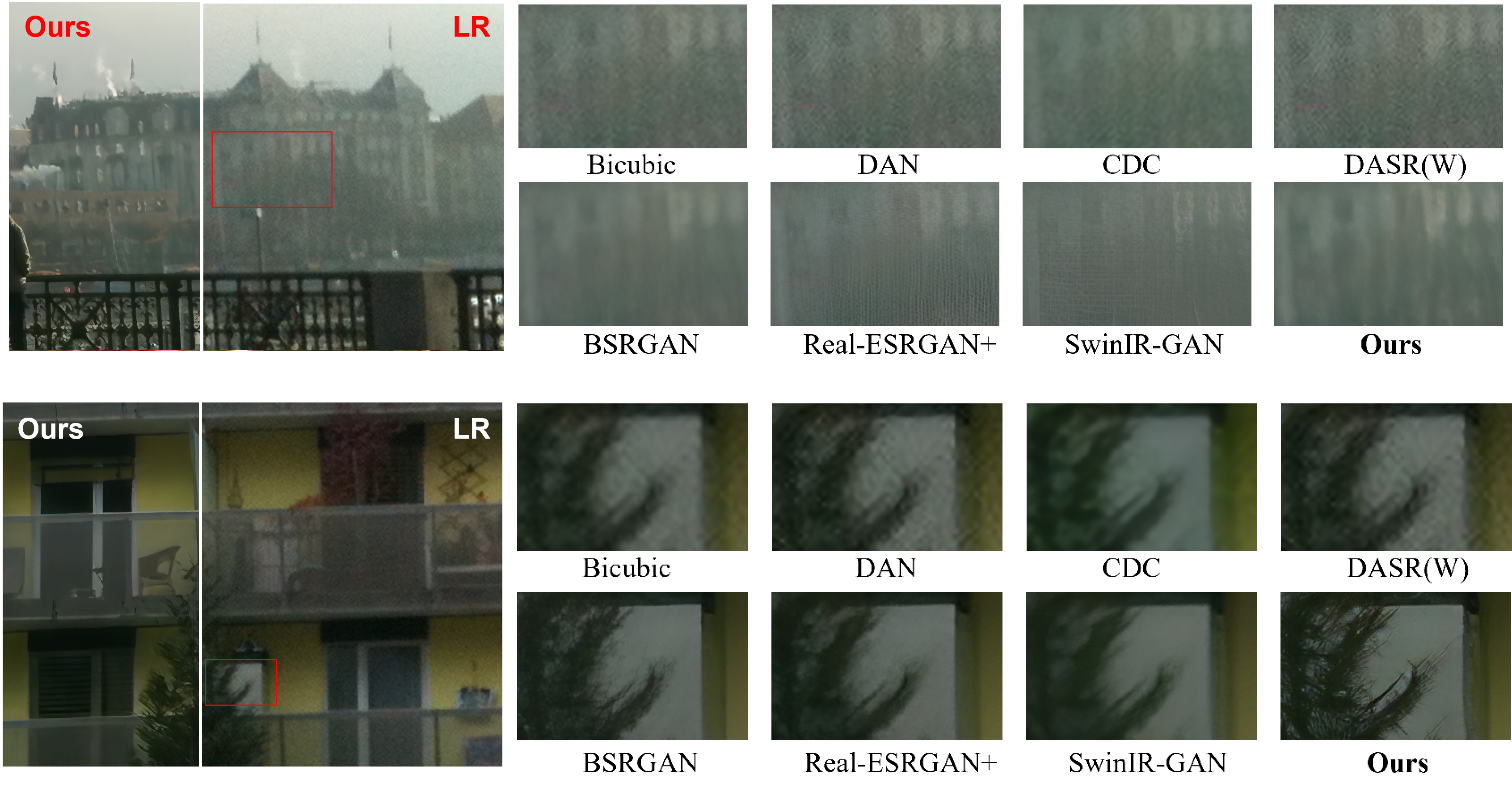}
        \caption{Visual comparisons on two real-world example with upscale factor 4. Our model can remove degradations and generate feasible details at the same time, while other GAN based methods tend to be either over-textured (first row) or over-smooth (second row). \red{Please zoom in for best view.}} \label{fig:real_results}
    \end{minipage}
\end{figure*}
\begin{table*}[t]
    \caption{Quantitative comparison with state-of-the-art methods on synthetic benchmarks. LR images are generated with a mixed degradation model of BSRGAN \cite{zhang2021designing} and Real-ESRGAN \cite{wang2021real}. PSNR/SSIM $\uparrow$: the higher, the better; LPIPS $\downarrow$: the lower, the better. LPIPS scores can better reflect texture quality, and the best and second performance are marked in \red{red} and \textcolor{blue}{blue}.} \label{tab:synthetic}
    \setlength{\tabcolsep}{1pt}
    \renewcommand*{\arraystretch}{1.2}
    \resizebox{\textwidth}{!}{
    \begin{tabular}{|c|c|cc|c|cc|c|cc|c|cc|c|cc|c|cc|c|}
    \hline
    \multirow{2}{*}{Method} & \multirow{2}{*}{Scale} & \multicolumn{3}{c|}{DIV2K Valid} & \multicolumn{3}{c|}{Set5} & \multicolumn{3}{c|}{Set14} & \multicolumn{3}{c|}{BSD100} & \multicolumn{3}{c|}{Urban100} & \multicolumn{3}{c|}{Manga109} \\ \cline{3-20} 
     &  & PSNR & SSIM & LPIPS & PSNR & SSIM & LPIPS & PSNR & SSIM & LPIPS & PSNR & SSIM & LPIPS & PSNR & SSIM & LPIPS & PSNR & SSIM & LPIPS \\ \hline
    CDC & $\times2$ & 24.93 & 0.6293 & 0.6588 & 25.35 & 0.6747 & 0.5153 & 22.74 & 0.5347 & 0.6229 & 23.64 & 0.5282 & 0.7073 & 20.94 & 0.5118 & 0.7001 & 21.60 & 0.6345 & 0.5723  \\ 
    DAN & $\times2$ & 24.69 & 0.5729 & 0.6219 & 25.27 & 0.6278 & 0.4658 & 22.79 & 0.5083 & 0.5639 & 23.46 & 0.4923 & 0.6384 & 20.93 & 0.4793 & 0.6603 & 21.78 & 0.5832 & 0.5639 \\ 
    DASR(W) & $\times2$ & 24.74 & 0.5767 & 0.6304 & 25.31 & 0.6312 & 0.4735 & 22.81 & 0.5110 & 0.5720 & 23.49 & 0.4958 & 0.6508 & 20.94 & 0.4819 & 0.6696 & 21.80 & 0.5878 & 0.5587 \\ \hdashline
    BSRGAN & $\times2$ & 26.60 & 0.7073 & 0.3182 & 27.65 & 0.7799 & \ublue{0.2027} & 24.59 & 0.6475 & \ublue{0.3013} & 24.88 & 0.5967 & 0.3769 & 22.76 & 0.6391 & 0.3199 & 24.64 & 0.7678 & 0.2285 \\ 
    Real-ESRGAN+ & $\times2$ & 25.50 & 0.6963 & \ublue{0.2993} & 26.73 & 0.7771 & 0.2157 & 23.65 & 0.6299 & 0.3023 & 24.11 & 0.5860 & \ublue{0.3433} & 21.66 & 0.6148 & \ublue{0.2876} & 23.88 & 0.7698 & \redb{0.2135} \\  
    SwinIR-GAN & $\times2$ & 25.33 & 0.6886 & 0.3313 & 27.07 & 0.7793 & 0.2093 & 23.76 & 0.6364 & 0.3128 & 23.83 & 0.5717 & 0.3707 & 21.54 & 0.6195 & 0.3003 & 23.56 & 0.7705 & 0.2283 \\ 
    \textbf{Ours} & $\times2$ & 25.26 & 0.6680 & \redb{0.2753} & 26.46 & 0.7470 & \redb{0.1964} & 23.38 & 0.5982 & \redb{0.2852} & 23.83 & 0.5599 & \redb{0.3264} & 21.90 & 0.5956 & \redb{0.2777} & 23.64 & 0.7407 & \ublue{0.2192} \\ \hline \hline
    CDC & $\times4$ & 23.11 & 0.5850 & 0.7132 & 19.99 & 0.5077 & 0.7168 & 20.38 & 0.4551 & 0.7377 & 21.75 & 0.4800 & 0.7707 & 19.42 & 0.4568 & 0.7345 & 19.92 & 0.5834 & 0.6102  \\ 
    DAN & $\times4$ & 24.22 & 0.5929 & 0.6881 & 20.85 & 0.5319 & 0.6771 & 21.44 & 0.4937 & 0.6758 & 22.52 & 0.4818 & 0.7438 & 20.20 & 0.4757 & 0.7228 & 21.02 & 0.5963 & 0.6198 \\ 
    DASR(W) & $\times4$ & 24.19 & 0.5920 & 0.7021 & 20.87 & 0.5336 & 0.6972 & 21.43 & 0.4953 & 0.6950 & 22.49 & 0.4818 & 0.7576 & 20.18 & 0.4752 & 0.7400 & 21.03 & 0.5975 & 0.6319 \\ \hdashline 
    BSRGAN & $\times4$ & 24.91 & 0.6500 & 0.3596 & 21.63 & 0.5573 & \redb{0.4683} & 22.17 & 0.5165 & \ublue{0.4173} & 22.95 & 0.5042 & \ublue{0.4405} & 20.91 & 0.5386 & 0.3874 & 22.45 & 0.6968 & \ublue{0.3039} \\ 
    Real-ESRGAN+ & $\times4$ & 23.80 & 0.6414 & 0.3696 & 21.31 & 0.5449 & 0.5068 & 21.54 & 0.5288 & 0.4271 & 22.43 & 0.5035 & 0.4693 & 19.90 & 0.5282 & 0.3838 & 21.97 & 0.6989 & 0.3073 \\ 
    SwinIR-GAN & $\times4$ & 24.13 & 0.6479 & \ublue{0.3543} & 20.91  & 0.5128 & 0.5115 & 21.58 & 0.5041 & 0.4487 & 22.23 & 0.4925 & 0.4447 & 20.01 & 0.5300 & \ublue{0.3592} & 22.21 & 0.7007 & 0.3044 \\ 
    \textbf{Ours} & $\times4$ & 23.77 & 0.6203 & \redb{0.3298} & 20.45 & 0.4863 & \ublue{0.4942} & 21.24 & 0.4809 & \redb{0.3801} & 22.11 & 0.4830 & \redb{0.4143} & 20.25 & 0.5243 & \redb{0.3566} & 21.98 & 0.6816 & \redb{0.2846} \\ \hline
    \end{tabular}
    }
\end{table*}
\begin{table*}[t]
\centering
\caption{Quantitative comparison with state-of-the-art methods on real-world benchmarks. NIQE $\downarrow$: the lower, the better. The best and second performance are marked in \red{red} and \textcolor{blue}{blue}. Some numbers of competitive methods are taken from \cite{wang2021real}.} \label{tab:real_benchmarks}
\resizebox{\textwidth}{!}{
\begin{tabular}{|c|ccccccccc|}
\hline
Datasets & Bicubic & DAN & RealSR & CDC & DASR(W) & BSRGAN & Real-ESRGAN+ & SwinIR-GAN & \textbf{Ours} \\ \hline
RealSR \cite{wang2021towards} & 6.2438 & 6.5673 & 6.8041 & 6.2376 & 8.1918 & 5.7355 & 4.7832 & \ublue{4.7644} & \redb{4.7434} \\ 
DRealSR \cite{wei2020cdc} & 6.5766 & 7.0720 & 7.7213 & 6.6359 & 9.1446 & 6.1362 & 4.8458 & \ublue{4.7053} & \redb{4.1987} \\ 
DPED-iphone \cite{ignatov2017dslr} & 6.0121 & 6.1414 & 5.5855 & 6.2738 & 6.9887 & 5.9906 & 5.2631 & \redb{4.9468} & \ublue{5.1066} \\ \hline
\end{tabular}
}
\end{table*}

In this experiment, we visualize the features in the codebook $\mathcal{Z}$ with pretrained $G$, which facilitates the understanding of the proposed framework by answering two questions: i) what priors are encoded in \hrp ii) how are they correlated to the semantics?

As shown in \cref{fig:code_vis}, we visualize the priors encoded in $\mathcal{Z}$ by projecting features to RGB pixel space with pretrained decoder $G$. In other words, we obtain the RGB patches of each vector $z_j \in \mathcal{Z}$ with $G(z_j)$, where the size of RGB patches is $8\times8$. 
Specifically, we explore how textures are encoded by single codes and combinations of different codes: 
\begin{itemize}
    \item \cref{fig:code_vis_single} shows that individual codes alone can represent some basic texture elements. However, when the same code is tiled onto a bigger feature map, \eg, $4\times4$, the decoder tends to preserve the color while producing a smooth image. This implies that a single code is not enough to represent complex textures. 
    \item \cref{fig:code_vis_multi} shows that complex and realistic textures can be generated by combining several different code samples, which indicates that the pretrained $\mathcal{Z}$ indeed learns to encode rich texture priors. 
    In addition, it can be observed that different combinations of code samples correspond to different semantics, such as, \cirnum{1} grass, \cirnum{2} plant and \cirnum{3} water.
    Please see the supplementary materials for more examples. 
\end{itemize}
Based on the above discussion, we conjecture that the individual codes in $\mathcal{Z}$ represent simple texture elements, while the diverse semantics are encoded in the combinations of multiple codes.

\subsection{Comparison with Existing Methods}

We compare the proposed FeMaSR with several state-of-the-art methods for blind SR, including CDC \cite{wei2020cdc}, DAN \cite{luo2020dan}, DASR(W) \cite{wang2021unsupervised}, RealSR \cite{ji2020realsr}, BSRGAN \cite{zhang2021designing}, Real-ESRGAN+ \cite{wang2021real} and SwinIR-GAN \cite{liang2021swinir}. Specifically, CDC proposed a divide-and-conquer architecture; DAN, DASR(W) and RealSR learned degradation models from LR inputs; BSRGAN, Real-ESRGAN+ and SwinIR-GAN used synthetic training data generated by handcrafted degradation models. We use the original codes and weights from the official public github repositories for all competing methods. Quantitative and qualitative results on both synthetic and real-world benchmarks are reported as follows.

\paragraph{Comparision on Synthetic Benchmarks}
As \cref{tab:synthetic} shows, our FeMaSR outperforms competing methods in LPIPS scores on most benchmarks (5 out of 6). 
Note that we focus on the LPIPS scores as it better captures the perceptual quality than other metrics (\eg, PSNR/SSIM) \cite{zhang2018perceptual,wang2021real,wang2021towards,zhang2021designing,yang2021gan}.
In addition, it can be observed that: in general, methods that learn the degradations, such as DAN and DASR(W), perform much worse than those using manually designed degradation models, which indicates the difficulties in learning complex real-world degradations. 
Furthermore, we compare the SR results qualitatively through visual inspection in \cref{fig:synthetic_results}.
It can be observed that in the first row, BSRGAN, Real-ESRGAN and SwinIR-GAN mistake the feather textures as noises and remove them. And in the second row, although the distortions are removed successfully, they all fail to generate feasible textures for the trees. In contrast, thanks to the semantic-aware HRP, our method does not have such problems and generates higher quality results. 

\paragraph{Comparison on Real-world Benchmarks} To make a fair comparison, we compare our method against state-of-the-art ones on three large real-world benchmarks and evaluate the results using a standard no-reference IQA metric NIQE. 
As \cref{tab:real_benchmarks} shows, our method outperforms competing methods in 2 out of 3 real-world benchmarks, which clearly demonstrates the effectiveness of our framework. 
In \cref{fig:real_results}, it can be observed that our FeMaSR produces sharp and clear textures without generating artifacts, while the other methods either fail to remove degradations or tend to be over-textured and over-smooth. Please see the supplementary materials for more results.   

\subsection{Ablation Study}

We conduct ablation experiments on four variations of our framework as shown in \cref{tab:ablation} to validate our design: Model-A, a baseline network by discarding Stage \RNum{1}, feature matching and residual shortcuts. It has a similar architecture with SwinIR, and is trained with GAN from scratch; Model-B, Model-A with pretrained decoder; Model-C, Model-B with pretrained codebook and feature matching; FeMaSR, full model with HRP and residual shortcuts; Model-D, FeMaSR based on \hrp without semantic guidance. 
\begin{figure}[t]
    \centering
    \newcommand{\hwidth}{1pt}
    \newcommand{\imgwidth}{0.24\linewidth}
    \newcommand{\patchwidth}{0.115\linewidth}
    \includegraphics[width=\linewidth]{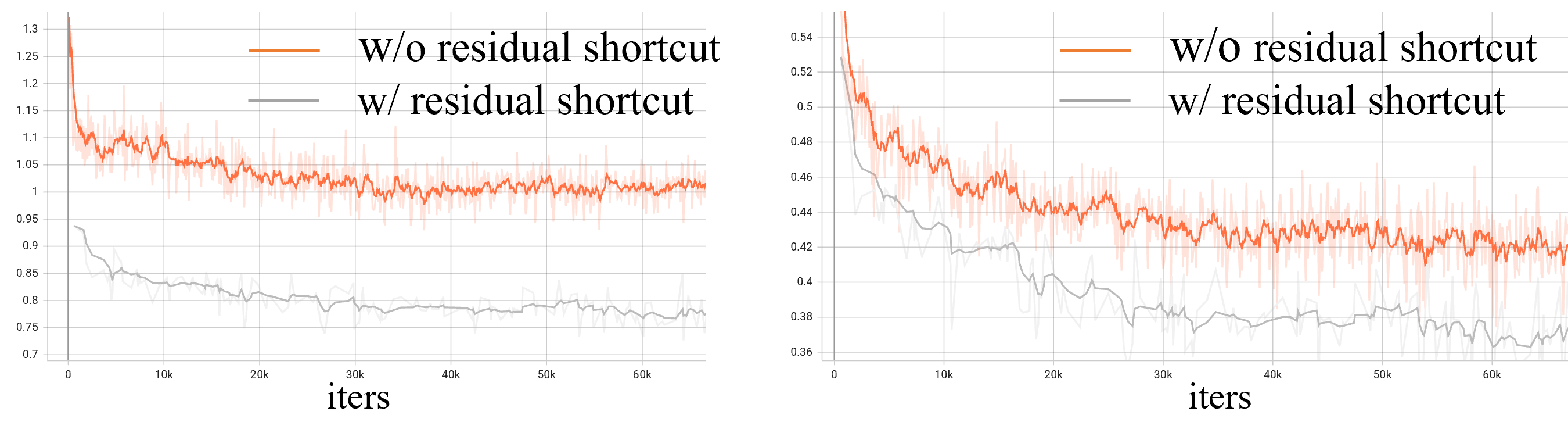} 
    \\
    \makebox[0.48\linewidth]{\small $\mathcal{L}_{fema}$}
    \makebox[0.48\linewidth]{\small $\mathcal{L}_{rec}$}
    \\
    \makebox[\linewidth]{\small (a) Stage \RNum{2} training loss curve w/ and w/o residual shortcut.} 
    \\[0.5em]
    \includegraphics[width=\linewidth]{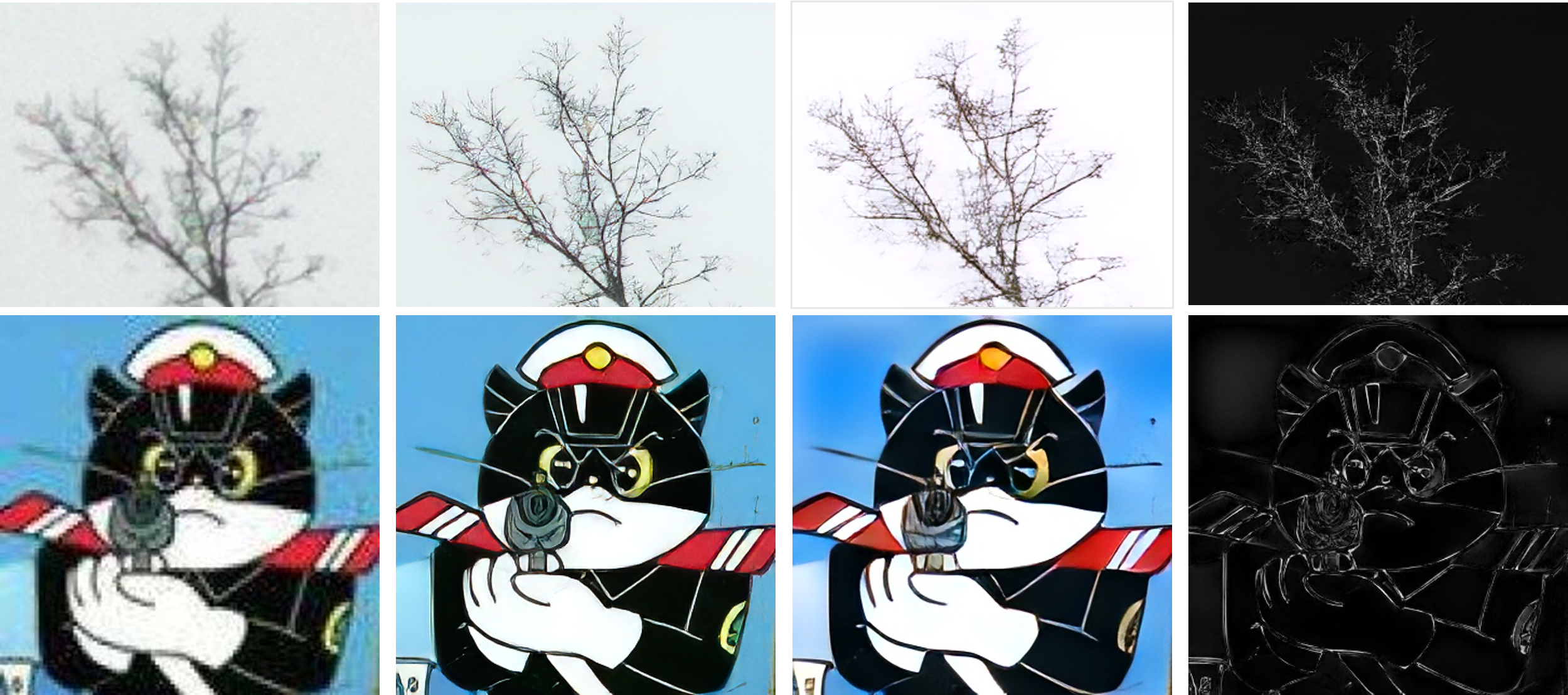}
    \\
    \makebox[\imgwidth]{\footnotesize LR input}
    \makebox[\imgwidth]{\footnotesize Full results}
    \makebox[\imgwidth]{\footnotesize Disable residual} 
    \makebox[\imgwidth]{\footnotesize Intensity difference} 
    \\
    \makebox[\linewidth]{\small (b) Results of disabling residual shortcut in test stage.} 
    \\
    \caption{Effectiveness of residual shortcut.} \label{fig:ablation_residual}
\end{figure}

\begin{table}[t]
\centering
\caption{Ablation study on synthetic benchmark DIV2K Valid with upscale factor of 2.} \label{tab:ablation}
\begin{tabular}{c|l|c}
\hline
Model ID & Model Variations &  LPIPS$\downarrow$ \\ \hline \hline
A & w/o \hrp & 0.3025 \\ \hline
B & + pretrained decoder & 0.2944 \\ \hline
C & \makecell{+ pretrained codebook \\ (with feature matching)} & 0.3358 \\ \hline
FeMaSR & + residual & \textbf{0.2753} \\ \hline \hline
D & FeMaSR w/o semantic & 0.2887 \\ \hline
\end{tabular}
\end{table}

\begin{figure}
    \centering
    \newcommand{\imgwidth}{0.24\linewidth}
    \includegraphics[width=\linewidth]{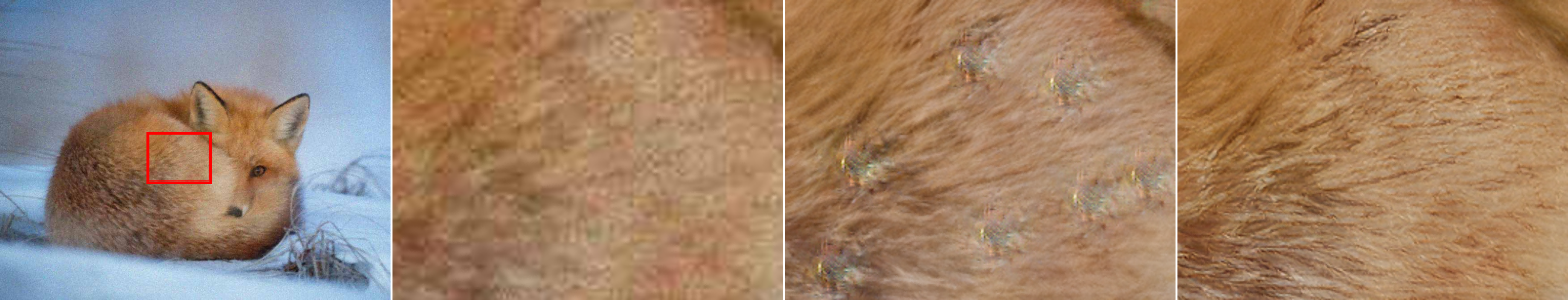}
    \\
    \makebox[\imgwidth]{\footnotesize LR input}
    \makebox[\imgwidth]{\footnotesize Bicubic ($\times 2$)}
    \makebox[\imgwidth]{\footnotesize Model-A} 
    \makebox[\imgwidth]{\footnotesize Model-B} 
    \\[0.2em]
    \includegraphics[width=\linewidth]{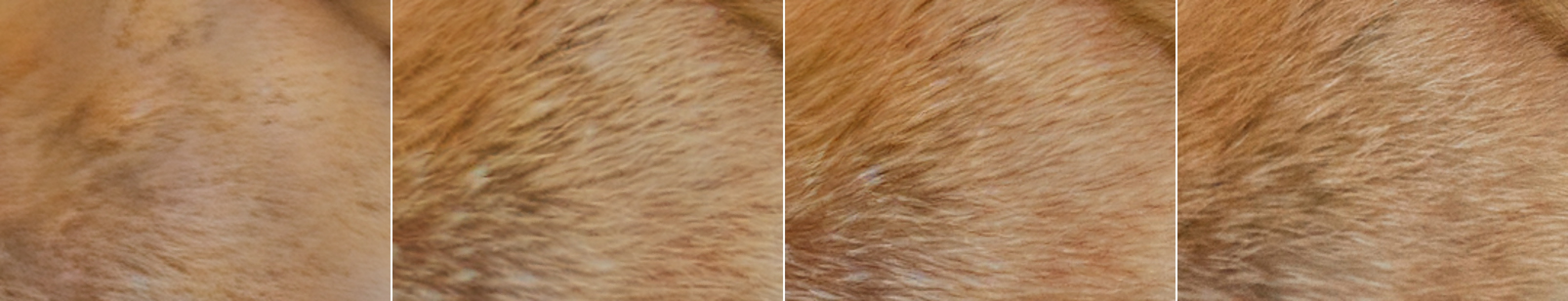}
    \\
    \makebox[\imgwidth]{\footnotesize Model-C}
    \makebox[\imgwidth]{\footnotesize Model-D}
    \makebox[\imgwidth]{\footnotesize FeMaSR} 
    \makebox[\imgwidth]{\footnotesize GT} 
    \\
    \caption{Visual examples of different model variations. \red{Please zoom in for best view.}}
    \label{fig:ablation}
\end{figure}

\paragraph{Effectiveness of Residual Shortcut} As claimed in \cref{sec:method}, residual shortcut helps optimization of feature matching process and complements possible matching errors. We verify them by removing the residual shortcut in training (Model-C) and testing stage respectively. As we can see in \cref{fig:ablation_residual}(a), the feature matching loss $\mathcal{L}_{fema}$ decreases much faster with residual shortcut. This indicates that residual shortcut is essential for the optimization of $\mathcal{L}_{fema}$. We can also observe a clear performance drop of model C without residual shortcut in \cref{tab:ablation} and \cref{fig:ablation}. We further demonstrate how residual shortcut helps to complement feature matching errors in \cref{fig:ablation_residual}(b). We can notice that model with disabled residual shortcut can already remove the distortions to a large extent. The residual shortcut mainly complements the color and edges. 

\paragraph{Effectiveness of \hrp} Model-[A, B and FeMaSR] validate the necessities of $\mathcal{Z}$ and $G$ in \hrp. As discussed above, the performance drop of Model-C is mainly due to the optimization difficulty brought by feature matching. Therefore, we do not use it to validate HRP. It can be observed that Model-B is better than Model-A since the pretrained decoder helps to stablize GAN training. However, both Model-A and Model-B cannot handle complex distortions without feature matching and tend to generate artifacts, see \cref{fig:ablation}. Meanwhile, the full model, FeMaSR, can make full use of HRP in both $G$ and $\mathcal{Z}$, and thereby has the best performance.   

\paragraph{Effectiveness of Semantic Guidance} We provide reconstruction training loss curve and LPIPS score in Stage \RNum{1} to show the benefits of semantic guidance. It can be seen that VQGAN with semantic guidance converges faster and performs better, resulting in a better HRP. This finally helps to improve the restoration performance, see \cref{tab:ablation} and \cref{fig:ablation}.  

\begin{figure}[t] \centering
    \includegraphics[width=0.49\linewidth]{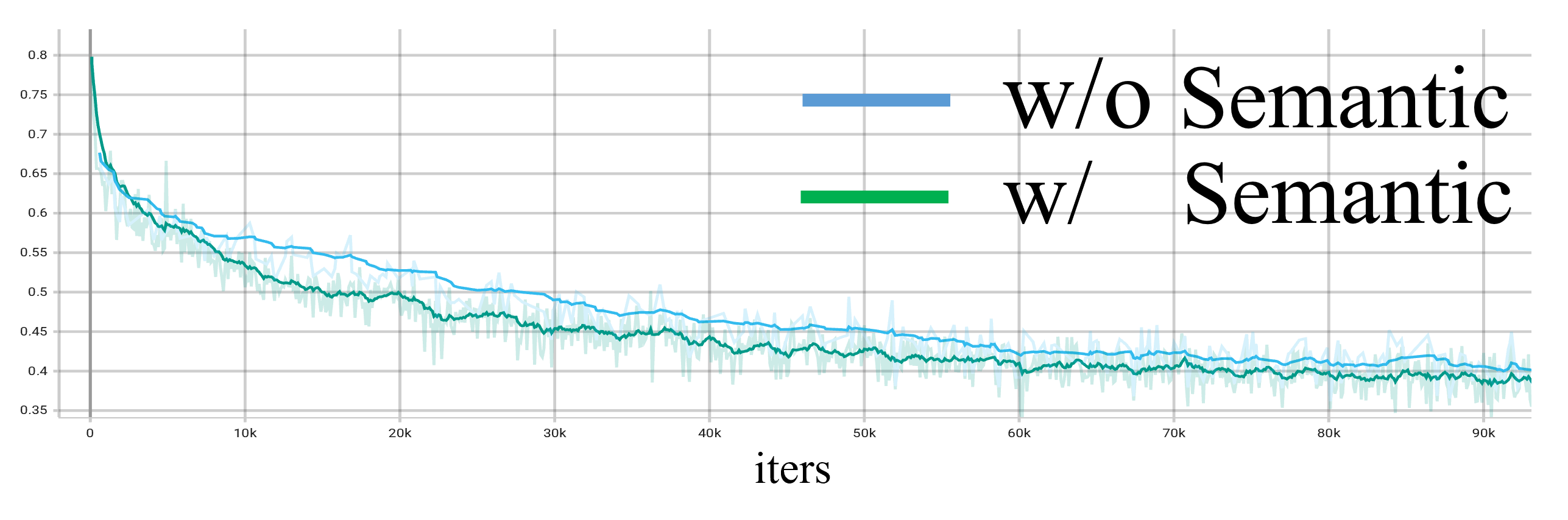}
    \raisebox{0.9\height}{
    \resizebox{0.49\linewidth}{!}{
    \begin{tabular}{lc}
        \toprule
        VQGAN & Reconstruction LPIPS$\downarrow$\\
        \midrule
        w/o semantic & 0.2032 \\
        w/ semantic &  0.1893 \\ 
        \bottomrule
    \end{tabular}
    }}
    \\
    \vspace{-0.2em}
    \makebox[0.49\linewidth]{\footnotesize (a) Stage \RNum{1} reconstruction loss curve} 
    \makebox[0.49\linewidth]{\footnotesize (b) Reconstruction quality on Div2k Valid}
    \\
    \caption{Effectiveness of semantic guidance.} 
    \label{tab:ablation_semantic}
    \vspace{-1.0em}
\end{figure}

\section{Conclusion}

In this paper, we have investigated the usage of implicit high-resolution priors (\hrp) encoded in the codebook and associated decoder of a pretrained VQGAN for real-world blind SR. 
In particular, we formulate the SR task to a feature matching problem between the LR features and distortion free HR feature codebook. Because \hrp is distortion free and fixed during SR stage, our FeMaSR is able to generate more realistic results with less artifacts than previous GAN based approaches. To train a better \hrp, we integrate semantic information to \hrp with features from pretrained VGG19 network. To facilitate optimization of feature matching loss, we introduce multi-scale residual shortcut connections to the pretrained decoder. Quantitative and qualitative experiments on both synthetic and real-world benchmarks demonstrate the superiority of the proposed FeMaSR for real-world LR images. 

\section{Acknowledgement}

This work was done at GAP Lab, CUHKSZ which is directed by Prof. Xiaoguang Han, and is supported by Alibaba Innovative Research, National Natural Science Foundation of China (62072383, 61702433, 62077039), the Fundamental Research Funds for the Central Universities (20720210044, 20720190006).
\bibliographystyle{ACM-Reference-Format}
\bibliography{qsr}


\begin{thebibliography}{65}


\ifx \showCODEN    \undefined \def \showCODEN     #1{\unskip}     \fi
\ifx \showDOI      \undefined \def \showDOI       #1{#1}\fi
\ifx \showISBNx    \undefined \def \showISBNx     #1{\unskip}     \fi
\ifx \showISBNxiii \undefined \def \showISBNxiii  #1{\unskip}     \fi
\ifx \showISSN     \undefined \def \showISSN      #1{\unskip}     \fi
\ifx \showLCCN     \undefined \def \showLCCN      #1{\unskip}     \fi
\ifx \shownote     \undefined \def \shownote      #1{#1}          \fi
\ifx \showarticletitle \undefined \def \showarticletitle #1{#1}   \fi
\ifx \showURL      \undefined \def \showURL       {\relax}        \fi
\providecommand\bibfield[2]{#2}
\providecommand\bibinfo[2]{#2}
\providecommand\natexlab[1]{#1}
\providecommand\showeprint[2][]{arXiv:#2}

\bibitem[Agustsson and Timofte(2017)]%
        {DIV2K}
\bibfield{author}{\bibinfo{person}{Eirikur Agustsson} {and}
  \bibinfo{person}{Radu Timofte}.} \bibinfo{year}{2017}\natexlab{}.
\newblock \showarticletitle{NTIRE 2017 Challenge on Single Image
  Super-Resolution: Dataset and Study}. In \bibinfo{booktitle}{\emph{CVPRW}}.
\newblock


\bibitem[Brock et~al\mbox{.}(2019)]%
        {brock2018biggan}
\bibfield{author}{\bibinfo{person}{Andrew Brock}, \bibinfo{person}{Jeff
  Donahue}, {and} \bibinfo{person}{Karen Simonyan}.}
  \bibinfo{year}{2019}\natexlab{}.
\newblock \showarticletitle{Large Scale {GAN} Training for High Fidelity
  Natural Image Synthesis}. In \bibinfo{booktitle}{\emph{ICLR}}.
\newblock


\bibitem[Chan et~al\mbox{.}(2021)]%
        {chan2021glean}
\bibfield{author}{\bibinfo{person}{Kelvin~CK Chan}, \bibinfo{person}{Xintao
  Wang}, \bibinfo{person}{Xiangyu Xu}, \bibinfo{person}{Jinwei Gu}, {and}
  \bibinfo{person}{Chen~Change Loy}.} \bibinfo{year}{2021}\natexlab{}.
\newblock \showarticletitle{Glean: Generative latent bank for large-factor
  image super-resolution}. In \bibinfo{booktitle}{\emph{CVPR}}.
  \bibinfo{pages}{14245--14254}.
\newblock


\bibitem[Chen et~al\mbox{.}(2020)]%
        {ChenSPARNet}
\bibfield{author}{\bibinfo{person}{Chaofeng Chen}, \bibinfo{person}{Dihong
  Gong}, \bibinfo{person}{Hao Wang}, \bibinfo{person}{Zhifeng Li}, {and}
  \bibinfo{person}{Kwan-Yee~K. Wong}.} \bibinfo{year}{2020}\natexlab{}.
\newblock \showarticletitle{Learning Spatial Attention for Face
  Super-Resolution}. In \bibinfo{booktitle}{\emph{IEEE TIP}}.
\newblock


\bibitem[Chen et~al\mbox{.}(2021a)]%
        {ChenPSFRGAN}
\bibfield{author}{\bibinfo{person}{Chaofeng Chen}, \bibinfo{person}{Xiaoming
  Li}, \bibinfo{person}{Yang Lingbo}, \bibinfo{person}{Xianhui Lin},
  \bibinfo{person}{Lei Zhang}, {and} \bibinfo{person}{KKY Wong}.}
  \bibinfo{year}{2021}\natexlab{a}.
\newblock \showarticletitle{Progressive Semantic-Aware Style Transformation for
  Blind Face Restoration}. In \bibinfo{booktitle}{\emph{CVPR}}.
\newblock


\bibitem[Chen et~al\mbox{.}(2021b)]%
        {chen2020IPT}
\bibfield{author}{\bibinfo{person}{Hanting Chen}, \bibinfo{person}{Yunhe Wang},
  \bibinfo{person}{Tianyu Guo}, \bibinfo{person}{Chang Xu},
  \bibinfo{person}{Yiping Deng}, \bibinfo{person}{Zhenhua Liu},
  \bibinfo{person}{Siwei Ma}, \bibinfo{person}{Chunjing Xu},
  \bibinfo{person}{Chao Xu}, {and} \bibinfo{person}{Wen Gao}.}
  \bibinfo{year}{2021}\natexlab{b}.
\newblock \showarticletitle{Pre-Trained Image Processing Transformer}. In
  \bibinfo{booktitle}{\emph{CVPR}}.
\newblock


\bibitem[Dai et~al\mbox{.}(2019)]%
        {dai2019SAN}
\bibfield{author}{\bibinfo{person}{Tao Dai}, \bibinfo{person}{Jianrui Cai},
  \bibinfo{person}{Yongbing Zhang}, \bibinfo{person}{Shu-Tao Xia}, {and}
  \bibinfo{person}{Lei Zhang}.} \bibinfo{year}{2019}\natexlab{}.
\newblock \showarticletitle{Second-order Attention Network for Single Image
  Super-Resolution}. In \bibinfo{booktitle}{\emph{CVPR}}.
  \bibinfo{pages}{11065--11074}.
\newblock


\bibitem[Dong et~al\mbox{.}(2014)]%
        {dong2014learning}
\bibfield{author}{\bibinfo{person}{Chao Dong}, \bibinfo{person}{Chen~Change
  Loy}, \bibinfo{person}{Kaiming He}, {and} \bibinfo{person}{Xiaoou Tang}.}
  \bibinfo{year}{2014}\natexlab{}.
\newblock \showarticletitle{Learning a deep convolutional network for image
  super-resolution}. In \bibinfo{booktitle}{\emph{ECCV}}. Springer,
  \bibinfo{pages}{184--199}.
\newblock


\bibitem[Esser et~al\mbox{.}(2021)]%
        {esser2021taming}
\bibfield{author}{\bibinfo{person}{Patrick Esser}, \bibinfo{person}{Robin
  Rombach}, {and} \bibinfo{person}{Bjorn Ommer}.}
  \bibinfo{year}{2021}\natexlab{}.
\newblock \showarticletitle{Taming transformers for high-resolution image
  synthesis}. In \bibinfo{booktitle}{\emph{CVPR}}.
  \bibinfo{pages}{12873--12883}.
\newblock


\bibitem[Fritsche et~al\mbox{.}(2019)]%
        {fritsche2019frequency}
\bibfield{author}{\bibinfo{person}{Manuel Fritsche}, \bibinfo{person}{Shuhang
  Gu}, {and} \bibinfo{person}{Radu Timofte}.} \bibinfo{year}{2019}\natexlab{}.
\newblock \showarticletitle{Frequency separation for real-world
  super-resolution}. In \bibinfo{booktitle}{\emph{ICCVW}}.
  \bibinfo{pages}{3599--3608}.
\newblock


\bibitem[Gatys et~al\mbox{.}(2015)]%
        {gatys2015texture}
\bibfield{author}{\bibinfo{person}{Leon Gatys}, \bibinfo{person}{Alexander~S
  Ecker}, {and} \bibinfo{person}{Matthias Bethge}.}
  \bibinfo{year}{2015}\natexlab{}.
\newblock \showarticletitle{Texture synthesis using convolutional neural
  networks}.
\newblock \bibinfo{journal}{\emph{NeurIPS}}  \bibinfo{volume}{28}
  (\bibinfo{year}{2015}), \bibinfo{pages}{262--270}.
\newblock


\bibitem[Gondal et~al\mbox{.}(2018)]%
        {gondal2018unreasonable}
\bibfield{author}{\bibinfo{person}{Muhammad~Waleed Gondal},
  \bibinfo{person}{Bernhard Sch{\"o}lkopf}, {and} \bibinfo{person}{Michael
  Hirsch}.} \bibinfo{year}{2018}\natexlab{}.
\newblock \showarticletitle{The unreasonable effectiveness of texture transfer
  for single image super-resolution}. In \bibinfo{booktitle}{\emph{ECCVW}}.
  Springer, \bibinfo{pages}{80--97}.
\newblock


\bibitem[Gu et~al\mbox{.}(2019a)]%
        {gu2019ikc}
\bibfield{author}{\bibinfo{person}{Jinjin Gu}, \bibinfo{person}{Hannan Lu},
  \bibinfo{person}{Wangmeng Zuo}, {and} \bibinfo{person}{Chao Dong}.}
  \bibinfo{year}{2019}\natexlab{a}.
\newblock \showarticletitle{Blind super-resolution with iterative kernel
  correction}. In \bibinfo{booktitle}{\emph{CVPR}}.
  \bibinfo{pages}{1604--1613}.
\newblock


\bibitem[Gu et~al\mbox{.}(2020)]%
        {gu2020image}
\bibfield{author}{\bibinfo{person}{Jinjin Gu}, \bibinfo{person}{Yujun Shen},
  {and} \bibinfo{person}{Bolei Zhou}.} \bibinfo{year}{2020}\natexlab{}.
\newblock \showarticletitle{Image processing using multi-code gan prior}. In
  \bibinfo{booktitle}{\emph{CVPR}}. \bibinfo{pages}{3012--3021}.
\newblock


\bibitem[Gu et~al\mbox{.}(2019b)]%
        {gu2019div8k}
\bibfield{author}{\bibinfo{person}{Shuhang Gu}, \bibinfo{person}{Andreas
  Lugmayr}, \bibinfo{person}{Martin Danelljan}, \bibinfo{person}{Manuel
  Fritsche}, \bibinfo{person}{Julien Lamour}, {and} \bibinfo{person}{Radu
  Timofte}.} \bibinfo{year}{2019}\natexlab{b}.
\newblock \showarticletitle{Div8k: Diverse 8k resolution image dataset}. In
  \bibinfo{booktitle}{\emph{ICCVW}}. IEEE, \bibinfo{pages}{3512--3516}.
\newblock


\bibitem[He et~al\mbox{.}(2016)]%
        {he2016deep}
\bibfield{author}{\bibinfo{person}{Kaiming He}, \bibinfo{person}{Xiangyu
  Zhang}, \bibinfo{person}{Shaoqing Ren}, {and} \bibinfo{person}{Jian Sun}.}
  \bibinfo{year}{2016}\natexlab{}.
\newblock \showarticletitle{Deep residual learning for image recognition}. In
  \bibinfo{booktitle}{\emph{CVPR}}. \bibinfo{pages}{770--778}.
\newblock


\bibitem[Huang et~al\mbox{.}(2017)]%
        {huang2017densely}
\bibfield{author}{\bibinfo{person}{Gao Huang}, \bibinfo{person}{Zhuang Liu},
  \bibinfo{person}{Laurens Van Der~Maaten}, {and} \bibinfo{person}{Kilian~Q
  Weinberger}.} \bibinfo{year}{2017}\natexlab{}.
\newblock \showarticletitle{Densely connected convolutional networks}. In
  \bibinfo{booktitle}{\emph{CVPR}}. \bibinfo{pages}{4700--4708}.
\newblock


\bibitem[Ignatov et~al\mbox{.}(2017)]%
        {ignatov2017dslr}
\bibfield{author}{\bibinfo{person}{Andrey Ignatov}, \bibinfo{person}{Nikolay
  Kobyshev}, \bibinfo{person}{Radu Timofte}, \bibinfo{person}{Kenneth Vanhoey},
  {and} \bibinfo{person}{Luc Van~Gool}.} \bibinfo{year}{2017}\natexlab{}.
\newblock \showarticletitle{Dslr-quality photos on mobile devices with deep
  convolutional networks}. In \bibinfo{booktitle}{\emph{ICCV}}.
  \bibinfo{pages}{3277--3285}.
\newblock


\bibitem[Ji et~al\mbox{.}(2020)]%
        {ji2020realsr}
\bibfield{author}{\bibinfo{person}{Xiaozhong Ji}, \bibinfo{person}{Yun Cao},
  \bibinfo{person}{Ying Tai}, \bibinfo{person}{Chengjie Wang},
  \bibinfo{person}{Jilin Li}, {and} \bibinfo{person}{Feiyue Huang}.}
  \bibinfo{year}{2020}\natexlab{}.
\newblock \showarticletitle{Real-world super-resolution via kernel estimation
  and noise injection}. In \bibinfo{booktitle}{\emph{CVPRW}}.
  \bibinfo{pages}{466--467}.
\newblock


\bibitem[Jiang et~al\mbox{.}(2021)]%
        {jiang2021c2match}
\bibfield{author}{\bibinfo{person}{Yuming Jiang}, \bibinfo{person}{Kelvin~CK
  Chan}, \bibinfo{person}{Xintao Wang}, \bibinfo{person}{Chen~Change Loy},
  {and} \bibinfo{person}{Ziwei Liu}.} \bibinfo{year}{2021}\natexlab{}.
\newblock \showarticletitle{Robust Reference-based Super-Resolution via
  C2-Matching}. In \bibinfo{booktitle}{\emph{CVPR}}.
  \bibinfo{pages}{2103--2112}.
\newblock


\bibitem[Karras et~al\mbox{.}(2018)]%
        {karras2017progressive}
\bibfield{author}{\bibinfo{person}{Tero Karras}, \bibinfo{person}{Timo Aila},
  \bibinfo{person}{Samuli Laine}, {and} \bibinfo{person}{Jaakko Lehtinen}.}
  \bibinfo{year}{2018}\natexlab{}.
\newblock \showarticletitle{Progressive growing of gans for improved quality,
  stability, and variation}.
\newblock \bibinfo{journal}{\emph{ICLR}} (\bibinfo{year}{2018}).
\newblock


\bibitem[Karras et~al\mbox{.}(2020)]%
        {karras2020analyzing}
\bibfield{author}{\bibinfo{person}{Tero Karras}, \bibinfo{person}{Samuli
  Laine}, \bibinfo{person}{Miika Aittala}, \bibinfo{person}{Janne Hellsten},
  \bibinfo{person}{Jaakko Lehtinen}, {and} \bibinfo{person}{Timo Aila}.}
  \bibinfo{year}{2020}\natexlab{}.
\newblock \showarticletitle{Analyzing and improving the image quality of
  stylegan}. In \bibinfo{booktitle}{\emph{CVPR}}. \bibinfo{pages}{8110--8119}.
\newblock


\bibitem[Kim et~al\mbox{.}(2016)]%
        {kim2016accurate}
\bibfield{author}{\bibinfo{person}{Jiwon Kim}, \bibinfo{person}{Jung~Kwon Lee},
  {and} \bibinfo{person}{Kyoung~Mu Lee}.} \bibinfo{year}{2016}\natexlab{}.
\newblock \showarticletitle{Accurate image super-resolution using very deep
  convolutional networks}. In \bibinfo{booktitle}{\emph{CVPR}}.
  \bibinfo{pages}{1646--1654}.
\newblock


\bibitem[Kingma and Ba(2014)]%
        {kingma2014adam}
\bibfield{author}{\bibinfo{person}{Diederik~P Kingma} {and}
  \bibinfo{person}{Jimmy Ba}.} \bibinfo{year}{2014}\natexlab{}.
\newblock \showarticletitle{Adam: A method for stochastic optimization}.
\newblock \bibinfo{journal}{\emph{arXiv preprint arXiv:1412.6980}}
  (\bibinfo{year}{2014}).
\newblock


\bibitem[Li et~al\mbox{.}(2020a)]%
        {li2020blind}
\bibfield{author}{\bibinfo{person}{Xiaoming Li}, \bibinfo{person}{Chaofeng
  Chen}, \bibinfo{person}{Shangchen Zhou}, \bibinfo{person}{Xianhui Lin},
  \bibinfo{person}{Wangmeng Zuo}, {and} \bibinfo{person}{Lei Zhang}.}
  \bibinfo{year}{2020}\natexlab{a}.
\newblock \showarticletitle{Blind face restoration via deep multi-scale
  component dictionaries}. In \bibinfo{booktitle}{\emph{ECCV}}. Springer,
  \bibinfo{pages}{399--415}.
\newblock


\bibitem[Li et~al\mbox{.}(2020b)]%
        {li2020enhanced}
\bibfield{author}{\bibinfo{person}{Xiaoming Li}, \bibinfo{person}{Wenyu Li},
  \bibinfo{person}{Dongwei Ren}, \bibinfo{person}{Hongzhi Zhang},
  \bibinfo{person}{Meng Wang}, {and} \bibinfo{person}{Wangmeng Zuo}.}
  \bibinfo{year}{2020}\natexlab{b}.
\newblock \showarticletitle{Enhanced blind face restoration with multi-exemplar
  images and adaptive spatial feature fusion}. In
  \bibinfo{booktitle}{\emph{CVPR}}. \bibinfo{pages}{2706--2715}.
\newblock


\bibitem[Li et~al\mbox{.}(2018)]%
        {li2018learning}
\bibfield{author}{\bibinfo{person}{Xiaoming Li}, \bibinfo{person}{Ming Liu},
  \bibinfo{person}{Yuting Ye}, \bibinfo{person}{Wangmeng Zuo},
  \bibinfo{person}{Liang Lin}, {and} \bibinfo{person}{Ruigang Yang}.}
  \bibinfo{year}{2018}\natexlab{}.
\newblock \showarticletitle{Learning warped guidance for blind face
  restoration}. In \bibinfo{booktitle}{\emph{ECCV}}. \bibinfo{pages}{272--289}.
\newblock


\bibitem[Liang et~al\mbox{.}(2021)]%
        {liang2021swinir}
\bibfield{author}{\bibinfo{person}{Jingyun Liang}, \bibinfo{person}{Jiezhang
  Cao}, \bibinfo{person}{Guolei Sun}, \bibinfo{person}{Kai Zhang},
  \bibinfo{person}{Luc Van~Gool}, {and} \bibinfo{person}{Radu Timofte}.}
  \bibinfo{year}{2021}\natexlab{}.
\newblock \showarticletitle{SwinIR: Image Restoration Using Swin Transformer}.
  In \bibinfo{booktitle}{\emph{ICCVW}}.
\newblock


\bibitem[Lim et~al\mbox{.}(2017)]%
        {lim2017enhanced}
\bibfield{author}{\bibinfo{person}{Bee Lim}, \bibinfo{person}{Sanghyun Son},
  \bibinfo{person}{Heewon Kim}, \bibinfo{person}{Seungjun Nah}, {and}
  \bibinfo{person}{Kyoung Mu~Lee}.} \bibinfo{year}{2017}\natexlab{}.
\newblock \showarticletitle{Enhanced deep residual networks for single image
  super-resolution}. In \bibinfo{booktitle}{\emph{CVPRW}}.
  \bibinfo{pages}{136--144}.
\newblock


\bibitem[Liu et~al\mbox{.}(2021)]%
        {liu2021swin}
\bibfield{author}{\bibinfo{person}{Ze Liu}, \bibinfo{person}{Yutong Lin},
  \bibinfo{person}{Yue Cao}, \bibinfo{person}{Han Hu}, \bibinfo{person}{Yixuan
  Wei}, \bibinfo{person}{Zheng Zhang}, \bibinfo{person}{Stephen Lin}, {and}
  \bibinfo{person}{Baining Guo}.} \bibinfo{year}{2021}\natexlab{}.
\newblock \showarticletitle{Swin transformer: Hierarchical vision transformer
  using shifted windows}.
\newblock \bibinfo{journal}{\emph{ICCV}} (\bibinfo{year}{2021}).
\newblock


\bibitem[Luo et~al\mbox{.}(2020)]%
        {luo2020dan}
\bibfield{author}{\bibinfo{person}{Zhengxiong Luo}, \bibinfo{person}{Yan
  Huang}, \bibinfo{person}{Shang Li}, \bibinfo{person}{Liang Wang}, {and}
  \bibinfo{person}{Tieniu Tan}.} \bibinfo{year}{2020}\natexlab{}.
\newblock \showarticletitle{Unfolding the Alternating Optimization for Blind
  Super Resolution}.
\newblock \bibinfo{journal}{\emph{NeurIPS}}  \bibinfo{volume}{33}
  (\bibinfo{year}{2020}).
\newblock


\bibitem[Maeda(2020)]%
        {maeda2020unpaired}
\bibfield{author}{\bibinfo{person}{Shunta Maeda}.}
  \bibinfo{year}{2020}\natexlab{}.
\newblock \showarticletitle{Unpaired image super-resolution using
  pseudo-supervision}. In \bibinfo{booktitle}{\emph{CVPR}}.
  \bibinfo{pages}{291--300}.
\newblock


\bibitem[Menon et~al\mbox{.}(2020)]%
        {menon2020pulse}
\bibfield{author}{\bibinfo{person}{Sachit Menon}, \bibinfo{person}{Alexandru
  Damian}, \bibinfo{person}{Shijia Hu}, \bibinfo{person}{Nikhil Ravi}, {and}
  \bibinfo{person}{Cynthia Rudin}.} \bibinfo{year}{2020}\natexlab{}.
\newblock \showarticletitle{Pulse: Self-supervised photo upsampling via latent
  space exploration of generative models}. In \bibinfo{booktitle}{\emph{CVPR}}.
  \bibinfo{pages}{2437--2445}.
\newblock


\bibitem[Miyato et~al\mbox{.}(2018)]%
        {miyato2018spectral}
\bibfield{author}{\bibinfo{person}{Takeru Miyato}, \bibinfo{person}{Toshiki
  Kataoka}, \bibinfo{person}{Masanori Koyama}, {and} \bibinfo{person}{Yuichi
  Yoshida}.} \bibinfo{year}{2018}\natexlab{}.
\newblock \showarticletitle{Spectral Normalization for Generative Adversarial
  Networks}. In \bibinfo{booktitle}{\emph{ICLR}}.
\newblock


\bibitem[Niu et~al\mbox{.}(2020)]%
        {niu2020HAN}
\bibfield{author}{\bibinfo{person}{Ben Niu}, \bibinfo{person}{Weilei Wen},
  \bibinfo{person}{Wenqi Ren}, \bibinfo{person}{Xiangde Zhang},
  \bibinfo{person}{Lianping Yang}, \bibinfo{person}{Shuzhen Wang},
  \bibinfo{person}{Kaihao Zhang}, \bibinfo{person}{Xiaochun Cao}, {and}
  \bibinfo{person}{Haifeng Shen}.} \bibinfo{year}{2020}\natexlab{}.
\newblock \showarticletitle{Single image super-resolution via a holistic
  attention network}. In \bibinfo{booktitle}{\emph{ECCV}}. Springer,
  \bibinfo{pages}{191--207}.
\newblock


\bibitem[Oord et~al\mbox{.}(2017)]%
        {vqvae}
\bibfield{author}{\bibinfo{person}{Aaron van~den Oord}, \bibinfo{person}{Oriol
  Vinyals}, {and} \bibinfo{person}{Koray Kavukcuoglu}.}
  \bibinfo{year}{2017}\natexlab{}.
\newblock \showarticletitle{Neural discrete representation learning}.
\newblock \bibinfo{journal}{\emph{NeurIPS}} (\bibinfo{year}{2017}).
\newblock


\bibitem[Pan et~al\mbox{.}(2020a)]%
        {pan2020dgp}
\bibfield{author}{\bibinfo{person}{Xingang Pan}, \bibinfo{person}{Xiaohang
  Zhan}, \bibinfo{person}{Bo Dai}, \bibinfo{person}{Dahua Lin},
  \bibinfo{person}{Chen~Change Loy}, {and} \bibinfo{person}{Ping Luo}.}
  \bibinfo{year}{2020}\natexlab{a}.
\newblock \showarticletitle{Exploiting Deep Generative Prior for Versatile
  Image Restoration and Manipulation}. In \bibinfo{booktitle}{\emph{ECCV}}.
\newblock


\bibitem[Pan et~al\mbox{.}(2020b)]%
        {pan2020exploiting}
\bibfield{author}{\bibinfo{person}{Xingang Pan}, \bibinfo{person}{Xiaohang
  Zhan}, \bibinfo{person}{Bo Dai}, \bibinfo{person}{Dahua Lin},
  \bibinfo{person}{Chen~Change Loy}, {and} \bibinfo{person}{Ping Luo}.}
  \bibinfo{year}{2020}\natexlab{b}.
\newblock \showarticletitle{Exploiting deep generative prior for versatile
  image restoration and manipulation}. In \bibinfo{booktitle}{\emph{ECCV}}.
  Springer, \bibinfo{pages}{262--277}.
\newblock


\bibitem[Paszke et~al\mbox{.}(2019)]%
        {pytorch}
\bibfield{author}{\bibinfo{person}{Adam Paszke}, \bibinfo{person}{Sam Gross},
  \bibinfo{person}{Francisco Massa}, \bibinfo{person}{Adam Lerer},
  \bibinfo{person}{James Bradbury}, \bibinfo{person}{Gregory Chanan},
  \bibinfo{person}{Trevor Killeen}, \bibinfo{person}{Zeming Lin},
  \bibinfo{person}{Natalia Gimelshein}, \bibinfo{person}{Luca Antiga},
  \bibinfo{person}{Alban Desmaison}, \bibinfo{person}{Andreas Kopf},
  \bibinfo{person}{Edward Yang}, \bibinfo{person}{Zachary DeVito},
  \bibinfo{person}{Martin Raison}, \bibinfo{person}{Alykhan Tejani},
  \bibinfo{person}{Sasank Chilamkurthy}, \bibinfo{person}{Benoit Steiner},
  \bibinfo{person}{Lu Fang}, \bibinfo{person}{Junjie Bai}, {and}
  \bibinfo{person}{Soumith Chintala}.} \bibinfo{year}{2019}\natexlab{}.
\newblock \showarticletitle{PyTorch: An Imperative Style, High-Performance Deep
  Learning Library}. In \bibinfo{booktitle}{\emph{NeurIPS}},
  Vol.~\bibinfo{volume}{32}. \bibinfo{pages}{8026--8037}.
\newblock


\bibitem[Razavi et~al\mbox{.}(2019)]%
        {vqvae2}
\bibfield{author}{\bibinfo{person}{Ali Razavi}, \bibinfo{person}{Aaron van~den
  Oord}, {and} \bibinfo{person}{Oriol Vinyals}.}
  \bibinfo{year}{2019}\natexlab{}.
\newblock \showarticletitle{Generating diverse high-fidelity images with
  vq-vae-2}. In \bibinfo{booktitle}{\emph{NeurIPS}}.
  \bibinfo{pages}{14866--14876}.
\newblock


\bibitem[Shocher et~al\mbox{.}(2018)]%
        {shocher2018zssr}
\bibfield{author}{\bibinfo{person}{Assaf Shocher}, \bibinfo{person}{Nadav
  Cohen}, {and} \bibinfo{person}{Michal Irani}.}
  \bibinfo{year}{2018}\natexlab{}.
\newblock \showarticletitle{“zero-shot” super-resolution using deep
  internal learning}. In \bibinfo{booktitle}{\emph{CVPR}}.
  \bibinfo{pages}{3118--3126}.
\newblock


\bibitem[Wan et~al\mbox{.}(2020)]%
        {wan2020bringing}
\bibfield{author}{\bibinfo{person}{Ziyu Wan}, \bibinfo{person}{Bo Zhang},
  \bibinfo{person}{Dongdong Chen}, \bibinfo{person}{Pan Zhang},
  \bibinfo{person}{Dong Chen}, \bibinfo{person}{Jing Liao}, {and}
  \bibinfo{person}{Fang Wen}.} \bibinfo{year}{2020}\natexlab{}.
\newblock \showarticletitle{Bringing old photos back to life}. In
  \bibinfo{booktitle}{\emph{CVPR}}. \bibinfo{pages}{2747--2757}.
\newblock


\bibitem[Wang et~al\mbox{.}(2021b)]%
        {wang2021unsupervised}
\bibfield{author}{\bibinfo{person}{Longguang Wang}, \bibinfo{person}{Yingqian
  Wang}, \bibinfo{person}{Xiaoyu Dong}, \bibinfo{person}{Qingyu Xu},
  \bibinfo{person}{Jungang Yang}, \bibinfo{person}{Wei An}, {and}
  \bibinfo{person}{Yulan Guo}.} \bibinfo{year}{2021}\natexlab{b}.
\newblock \showarticletitle{Unsupervised Degradation Representation Learning
  for Blind Super-Resolution}. In \bibinfo{booktitle}{\emph{CVPR}}.
  \bibinfo{pages}{10581--10590}.
\newblock


\bibitem[Wang et~al\mbox{.}(2021a)]%
        {wang2021towards}
\bibfield{author}{\bibinfo{person}{Xintao Wang}, \bibinfo{person}{Yu Li},
  \bibinfo{person}{Honglun Zhang}, {and} \bibinfo{person}{Ying Shan}.}
  \bibinfo{year}{2021}\natexlab{a}.
\newblock \showarticletitle{Towards Real-World Blind Face Restoration with
  Generative Facial Prior}. In \bibinfo{booktitle}{\emph{CVPR}}.
  \bibinfo{pages}{9168--9178}.
\newblock


\bibitem[Wang et~al\mbox{.}(2021c)]%
        {wang2021real}
\bibfield{author}{\bibinfo{person}{Xintao Wang}, \bibinfo{person}{Liangbin
  Xie}, \bibinfo{person}{Chao Dong}, {and} \bibinfo{person}{Ying Shan}.}
  \bibinfo{year}{2021}\natexlab{c}.
\newblock \showarticletitle{Real-ESRGAN: Training Real-World Blind
  Super-Resolution with Pure Synthetic Data}.
\newblock \bibinfo{journal}{\emph{ICCVW}} (\bibinfo{year}{2021}).
\newblock


\bibitem[Wang et~al\mbox{.}(2018)]%
        {wang2018sftgan}
\bibfield{author}{\bibinfo{person}{Xintao Wang}, \bibinfo{person}{Ke Yu},
  \bibinfo{person}{Chao Dong}, {and} \bibinfo{person}{Chen~Change Loy}.}
  \bibinfo{year}{2018}\natexlab{}.
\newblock \showarticletitle{Recovering realistic texture in image
  super-resolution by deep spatial feature transform}. In
  \bibinfo{booktitle}{\emph{CVPR}}.
\newblock


\bibitem[Wei et~al\mbox{.}(2020)]%
        {wei2020cdc}
\bibfield{author}{\bibinfo{person}{Pengxu Wei}, \bibinfo{person}{Ziwei Xie},
  \bibinfo{person}{Hannan Lu}, \bibinfo{person}{Zongyuan Zhan},
  \bibinfo{person}{Qixiang Ye}, \bibinfo{person}{Wangmeng Zuo}, {and}
  \bibinfo{person}{Liang Lin}.} \bibinfo{year}{2020}\natexlab{}.
\newblock \showarticletitle{Component divide-and-conquer for real-world image
  super-resolution}. In \bibinfo{booktitle}{\emph{ECCV}}. Springer,
  \bibinfo{pages}{101--117}.
\newblock


\bibitem[Wei et~al\mbox{.}(2021)]%
        {wei2021unsupervised}
\bibfield{author}{\bibinfo{person}{Yunxuan Wei}, \bibinfo{person}{Shuhang Gu},
  \bibinfo{person}{Yawei Li}, \bibinfo{person}{Radu Timofte},
  \bibinfo{person}{Longcun Jin}, {and} \bibinfo{person}{Hengjie Song}.}
  \bibinfo{year}{2021}\natexlab{}.
\newblock \showarticletitle{Unsupervised real-world image super resolution via
  domain-distance aware training}. In \bibinfo{booktitle}{\emph{CVPR}}.
  \bibinfo{pages}{13385--13394}.
\newblock


\bibitem[Yang et~al\mbox{.}(2020)]%
        {yang2020ttsr}
\bibfield{author}{\bibinfo{person}{Fuzhi Yang}, \bibinfo{person}{Huan Yang},
  \bibinfo{person}{Jianlong Fu}, \bibinfo{person}{Hongtao Lu}, {and}
  \bibinfo{person}{Baining Guo}.} \bibinfo{year}{2020}\natexlab{}.
\newblock \showarticletitle{Learning Texture Transformer Network for Image
  Super-Resolution}. In \bibinfo{booktitle}{\emph{CVPR}}.
\newblock


\bibitem[Yang et~al\mbox{.}(2021)]%
        {yang2021gan}
\bibfield{author}{\bibinfo{person}{Tao Yang}, \bibinfo{person}{Peiran Ren},
  \bibinfo{person}{Xuansong Xie}, {and} \bibinfo{person}{Lei Zhang}.}
  \bibinfo{year}{2021}\natexlab{}.
\newblock \showarticletitle{GAN Prior Embedded Network for Blind Face
  Restoration in the Wild}. In \bibinfo{booktitle}{\emph{CVPR}}.
  \bibinfo{pages}{672--681}.
\newblock


\bibitem[Zhang et~al\mbox{.}(2021b)]%
        {zhang2021blind}
\bibfield{author}{\bibinfo{person}{Jiahui Zhang}, \bibinfo{person}{Shijian Lu},
  \bibinfo{person}{Fangneng Zhan}, {and} \bibinfo{person}{Yingchen Yu}.}
  \bibinfo{year}{2021}\natexlab{b}.
\newblock \showarticletitle{Blind Image Super-Resolution via Contrastive
  Representation Learning}.
\newblock \bibinfo{journal}{\emph{arXiv preprint arXiv:2107.00708}}
  (\bibinfo{year}{2021}).
\newblock


\bibitem[Zhang et~al\mbox{.}(2020)]%
        {zhang2020deep}
\bibfield{author}{\bibinfo{person}{Kai Zhang}, \bibinfo{person}{Luc~Van Gool},
  {and} \bibinfo{person}{Radu Timofte}.} \bibinfo{year}{2020}\natexlab{}.
\newblock \showarticletitle{Deep unfolding network for image super-resolution}.
  In \bibinfo{booktitle}{\emph{CVPR}}. \bibinfo{pages}{3217--3226}.
\newblock


\bibitem[Zhang et~al\mbox{.}(2021a)]%
        {zhang2021designing}
\bibfield{author}{\bibinfo{person}{Kai Zhang}, \bibinfo{person}{Jingyun Liang},
  \bibinfo{person}{Luc Van~Gool}, {and} \bibinfo{person}{Radu Timofte}.}
  \bibinfo{year}{2021}\natexlab{a}.
\newblock \showarticletitle{Designing a practical degradation model for deep
  blind image super-resolution}.
\newblock \bibinfo{journal}{\emph{ICCV}} (\bibinfo{year}{2021}).
\newblock


\bibitem[Zhang et~al\mbox{.}(2018d)]%
        {zhang2018learning}
\bibfield{author}{\bibinfo{person}{Kai Zhang}, \bibinfo{person}{Wangmeng Zuo},
  {and} \bibinfo{person}{Lei Zhang}.} \bibinfo{year}{2018}\natexlab{d}.
\newblock \showarticletitle{Learning a single convolutional super-resolution
  network for multiple degradations}. In \bibinfo{booktitle}{\emph{CVPR}}.
  \bibinfo{pages}{3262--3271}.
\newblock


\bibitem[Zhang et~al\mbox{.}(2019d)]%
        {zhang2019deep}
\bibfield{author}{\bibinfo{person}{Kai Zhang}, \bibinfo{person}{Wangmeng Zuo},
  {and} \bibinfo{person}{Lei Zhang}.} \bibinfo{year}{2019}\natexlab{d}.
\newblock \showarticletitle{Deep plug-and-play super-resolution for arbitrary
  blur kernels}. In \bibinfo{booktitle}{\emph{CVPR}}.
  \bibinfo{pages}{1671--1681}.
\newblock


\bibitem[Zhang et~al\mbox{.}(2018a)]%
        {zhang2018perceptual}
\bibfield{author}{\bibinfo{person}{Richard Zhang}, \bibinfo{person}{Phillip
  Isola}, \bibinfo{person}{Alexei~A Efros}, \bibinfo{person}{Eli Shechtman},
  {and} \bibinfo{person}{Oliver Wang}.} \bibinfo{year}{2018}\natexlab{a}.
\newblock \showarticletitle{The Unreasonable Effectiveness of Deep Features as
  a Perceptual Metric}. In \bibinfo{booktitle}{\emph{CVPR}}.
\newblock


\bibitem[Zhang et~al\mbox{.}(2019b)]%
        {zhang2019ranksrgan}
\bibfield{author}{\bibinfo{person}{Wenlong Zhang}, \bibinfo{person}{Yihao Liu},
  \bibinfo{person}{Chao Dong}, {and} \bibinfo{person}{Yu Qiao}.}
  \bibinfo{year}{2019}\natexlab{b}.
\newblock \showarticletitle{Ranksrgan: Generative adversarial networks with
  ranker for image super-resolution}. In \bibinfo{booktitle}{\emph{ICCV}}.
  \bibinfo{pages}{3096--3105}.
\newblock


\bibitem[Zhang et~al\mbox{.}(2018b)]%
        {zhang2018image}
\bibfield{author}{\bibinfo{person}{Yulun Zhang}, \bibinfo{person}{Kunpeng Li},
  \bibinfo{person}{Kai Li}, \bibinfo{person}{Lichen Wang},
  \bibinfo{person}{Bineng Zhong}, {and} \bibinfo{person}{Yun Fu}.}
  \bibinfo{year}{2018}\natexlab{b}.
\newblock \showarticletitle{Image super-resolution using very deep residual
  channel attention networks}. In \bibinfo{booktitle}{\emph{ECCV}}.
  \bibinfo{pages}{286--301}.
\newblock


\bibitem[Zhang et~al\mbox{.}(2019a)]%
        {zhang2019rnan}
\bibfield{author}{\bibinfo{person}{Yulun Zhang}, \bibinfo{person}{Kunpeng Li},
  \bibinfo{person}{Kai Li}, \bibinfo{person}{Bineng Zhong}, {and}
  \bibinfo{person}{Yun Fu}.} \bibinfo{year}{2019}\natexlab{a}.
\newblock \showarticletitle{Residual Non-local Attention Networks for Image
  Restoration}. In \bibinfo{booktitle}{\emph{ICLR}}.
\newblock


\bibitem[Zhang et~al\mbox{.}(2018c)]%
        {zhang2018residual}
\bibfield{author}{\bibinfo{person}{Yulun Zhang}, \bibinfo{person}{Yapeng Tian},
  \bibinfo{person}{Yu Kong}, \bibinfo{person}{Bineng Zhong}, {and}
  \bibinfo{person}{Yun Fu}.} \bibinfo{year}{2018}\natexlab{c}.
\newblock \showarticletitle{Residual dense network for image super-resolution}.
  In \bibinfo{booktitle}{\emph{CVPR}}. \bibinfo{pages}{2472--2481}.
\newblock


\bibitem[Zhang et~al\mbox{.}(2019c)]%
        {zhang2019image}
\bibfield{author}{\bibinfo{person}{Zhifei Zhang}, \bibinfo{person}{Zhaowen
  Wang}, \bibinfo{person}{Zhe Lin}, {and} \bibinfo{person}{Hairong Qi}.}
  \bibinfo{year}{2019}\natexlab{c}.
\newblock \showarticletitle{Image super-resolution by neural texture transfer}.
  In \bibinfo{booktitle}{\emph{CVPR}}. \bibinfo{pages}{7982--7991}.
\newblock


\bibitem[Zheng et~al\mbox{.}(2018)]%
        {zheng2018crossnet}
\bibfield{author}{\bibinfo{person}{Haitian Zheng}, \bibinfo{person}{Mengqi Ji},
  \bibinfo{person}{Haoqian Wang}, \bibinfo{person}{Yebin Liu}, {and}
  \bibinfo{person}{Lu Fang}.} \bibinfo{year}{2018}\natexlab{}.
\newblock \showarticletitle{CrossNet: An End-to-end Reference-based Super
  Resolution Network using Cross-scale Warping}. In
  \bibinfo{booktitle}{\emph{ECCV}}. \bibinfo{pages}{88--104}.
\newblock


\bibitem[Zhou and Susstrunk(2019)]%
        {zhou2019kmsr}
\bibfield{author}{\bibinfo{person}{Ruofan Zhou} {and} \bibinfo{person}{Sabine
  Susstrunk}.} \bibinfo{year}{2019}\natexlab{}.
\newblock \showarticletitle{Kernel modeling super-resolution on real
  low-resolution images}. In \bibinfo{booktitle}{\emph{CVPR}}.
  \bibinfo{pages}{2433--2443}.
\newblock


\bibitem[Zhou et~al\mbox{.}(2020)]%
        {zhou2020cross}
\bibfield{author}{\bibinfo{person}{Shangchen Zhou}, \bibinfo{person}{Jiawei
  Zhang}, \bibinfo{person}{Wangmeng Zuo}, {and} \bibinfo{person}{Chen~Change
  Loy}.} \bibinfo{year}{2020}\natexlab{}.
\newblock \showarticletitle{Cross-Scale Internal Graph Neural Network for Image
  Super-Resolution}. In \bibinfo{booktitle}{\emph{NeurIPS}}.
\newblock


\bibitem[Zhu et~al\mbox{.}(2017)]%
        {zhu2017unpaired}
\bibfield{author}{\bibinfo{person}{Jun-Yan Zhu}, \bibinfo{person}{Taesung
  Park}, \bibinfo{person}{Phillip Isola}, {and} \bibinfo{person}{Alexei~A
  Efros}.} \bibinfo{year}{2017}\natexlab{}.
\newblock \showarticletitle{Unpaired image-to-image translation using
  cycle-consistent adversarial networks}. In \bibinfo{booktitle}{\emph{ICCV}}.
  \bibinfo{pages}{2223--2232}.
\newblock


\end{thebibliography}

\clearpage

\appendix

\section{More Implementation Details} 

\subsection{Network Architectures of VQGAN} \label{sec:network_details}

\begin{figure*}[h]
    \centering
    \includegraphics[width=1.0\linewidth]{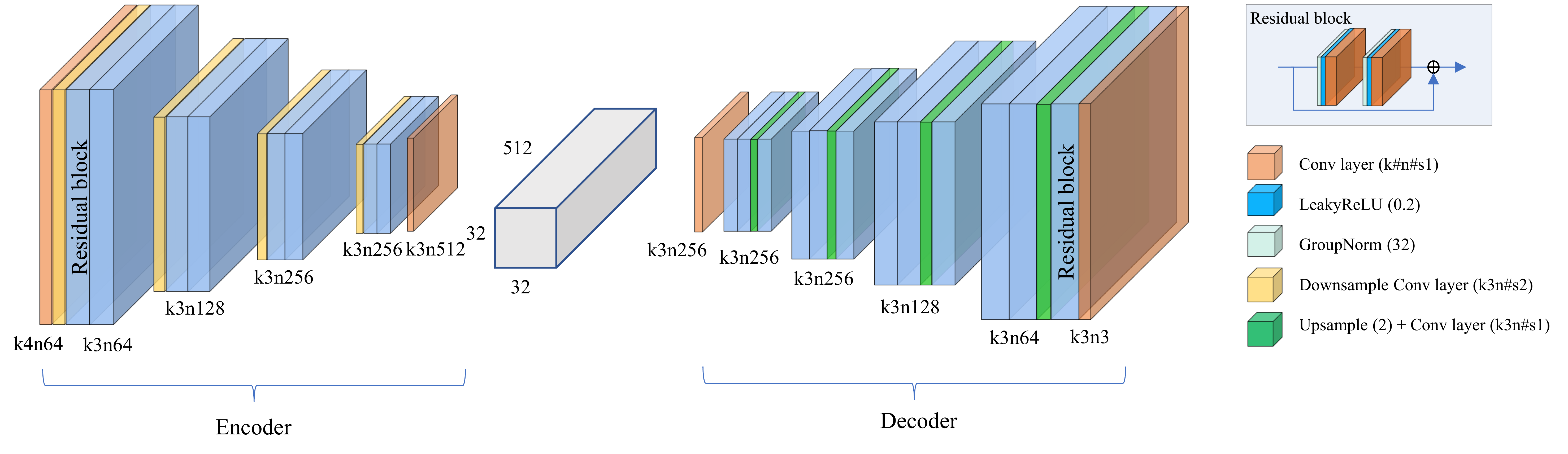}
    \caption{Details of hyper parameters for VQGAN. ``k3n64s1'' denotes single convolution layer with kernel size $3\times3$, output channel $64$ and stride size $1$, ``\#'' means the corresponding value. We use group normalization with $32$ group numbers, leakyrelu with $0.2$ as negative slope. The upsample convolution is a nearest upsample layer with scale factor 2 followed by a convolution layer.}
    \label{fig:network_params}
\end{figure*}

In complementary to the simple network architecture in the paper, we provide details of hyper parameters for the VQGAN encoder and decoder in \cref{fig:network_params}. We use the codebook $\bm{f}^{tp}$ with size $1024\times512$ in our experiment. The input image is of size $256\times256$ and downsampled into $32\times32$ feature maps. Convolution layers with ``k3n\#s1'' are used to match feature channels with the codebook feature dimensions before and after the feature quantization process.

We perform experiments to select suitable codebook numbers for our model, see \cref{tab:codebook_num} for the results. It can be observed that more codes generally leads to better reconstruction performance, but the improvements is marginal when the number is sufficiently large. We empirically select 1024 as a good balance between performance and computation cost. 

\begin{table}[h]
    \caption{VQGAN reconstruction performance with different code numbers.}
    \label{tab:codebook_num}
    \centering
    \begin{tabular}{c|cccc}
    \hline
     Codebook number & 256 & 512 & 1024 & 2048 \\ \hline
     Reconstruction LPIPS$\downarrow$ & 0.2165 & 0.2044 & 0.1893 & 0.1837 \\
    \hline 
    \end{tabular}
\end{table}

\subsection{Details of Synthetic Dataset} \label{sec:data_details}

\begin{figure*}[h]
    \centering
    \includegraphics[width=1.0\linewidth]{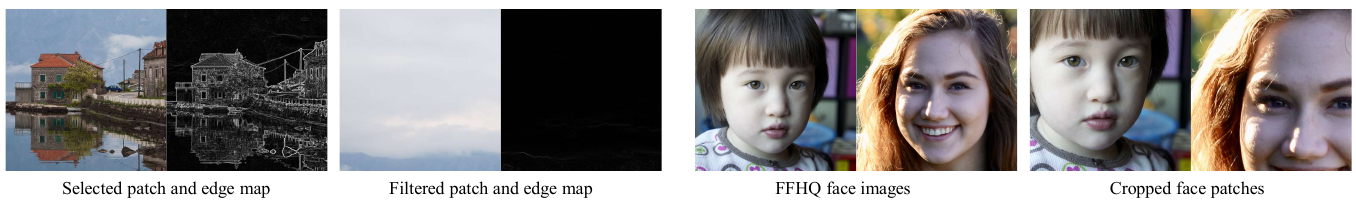}
    \caption{Examples of cropped training HR patches.}
    \label{fig:train_patch}
\end{figure*}

\paragraph{Generation of training HR patches.} As described in the paper, we use high resolution images from DIV2K \cite{DIV2K}, Flickr2K \cite{lim2017enhanced}, DIV8K \cite{gu2019div8k} and \num{10000} face images from FFHQ \cite{karras2017progressive} to generate HR training patches of size $512\times512$. The overall summary of training images are shown in \cref{tab:train_patch}, and some examples are shown in \cref{fig:train_patch}. Details to obtain the patches are as follow:

For the first three datasets which contains natural images, we crop the patches with the following steps:
\begin{enumerate}
    \item Crop non-overlapping $512\times512$ patches; 
    \item Filter patches with few textures (or edges). For this purpose, we first calculate the sobel edge map of the patch, then compute mean and variation of the edge map, denoted as $\mu, \sigma^2$. Because edge map is sparse, more edges means bigger $\sigma^2$ and $\mu$, we therefore empirically filter patches whose $\sigma^2 < 10$;
\end{enumerate}

For the FFHQ face dataset, the images are well-aligned faces of size $1024\times1024$. Previous non-overlap cropping would cause content bias for this dataset. Therefore, we first randomly resize the image with scale factor between $[0.5, 1.0]$, and then random crop only one patch from each image.  

\begin{table*}[h]
    \centering
    \caption{Details of HR training datasets.} \label{tab:train_patch}
    \begin{tabular}{l|cccc|r}
    \hline
       Dataset  &  DIV2K & DIV8K & Flickr2K & FFHQ & Total \\ \hline
       Number of full image  & \num{800} & \num{1500} & \num{2550} & \num{10000} & \num{14850} \\
       Typical image size $W\times H$ & $2032\times1344$ & $6720\times3840$ & $2032\times1344$ & $1024\times1024$ & - \\
       Number of cropped patches & \num{8257} & \num{90892} & \num{27056} & \num{10000} & \num{136205} \\
    \hline
    \end{tabular}
    \label{tab:my_label}
\end{table*}

\paragraph{Online generation of training pairs.} We use the same degradation model as BSRGAN~\cite{zhang2021designing} to generate corresponding LR images online. To be specific, the input HR patches are first randomly cropped to $256\times256$ patches, and then degraded with \texttt{degradation\_bsrgan}\footnote{\label{fn_bsrgan}\url{https://github.com/cszn/BSRGAN/blob/main/utils/utils_blindsr.py}} function with scale factors 2 and 4 to generate training pairs. 

\paragraph{Generation of synthetic testing benchmarks.} For a fair comparison, we use a mixed degradation model of BSRGAN and Real-ESRGAN to synthesize testing LR images. Specifically, we use \texttt{degradation\_bsrgan\_plus}\footnote{\label{fn_bsrgan}\url{https://github.com/cszn/BSRGAN/blob/main/utils/utils_blindsr.py}} function with scale factor 2 and 4 to generate testing pairs with a fixed random seed 123. 

\section{More Results}

\begin{figure*}[t]
    \centering
    \includegraphics[width=1.0\linewidth]{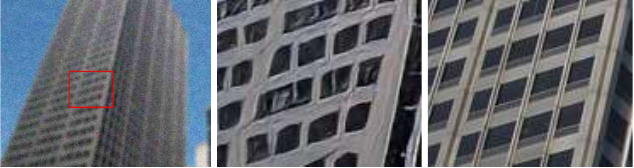}
    \\
    \makebox[0.32\linewidth]{LR Input}
    \makebox[0.32\linewidth]{FeMaSR}
    \makebox[0.32\linewidth]{Ground Truth}
    \caption{Failure case: a building image with straight lines.} \label{fig:limitations}
\end{figure*}

\subsection{Visualization of HRP} \label{sec:vis_codebook}

\begin{figure*}
    \centering
    \includegraphics[width=1.\textwidth]{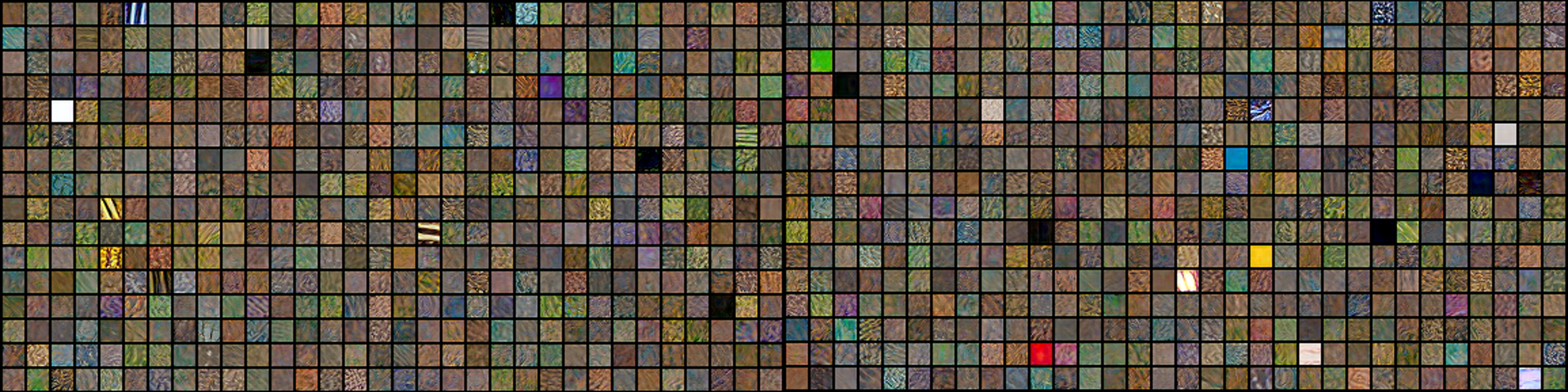}
    \caption{Visualization overview of all the 1024 codes in HRP.}
    \label{fig:codebook_overview}
\end{figure*}

\begin{figure*}
    \centering
    \newcommand{\namewidth}{0.1\textwidth}
    \newcommand{\nameheight}{2.2\height}
    \makebox[0.01\linewidth]{}
    \makebox[0.44\linewidth]{Original images}
    \makebox[0.44\linewidth]{Reconstructed images with VQGAN} 
    \\
    \includegraphics[width=1.0\textwidth]{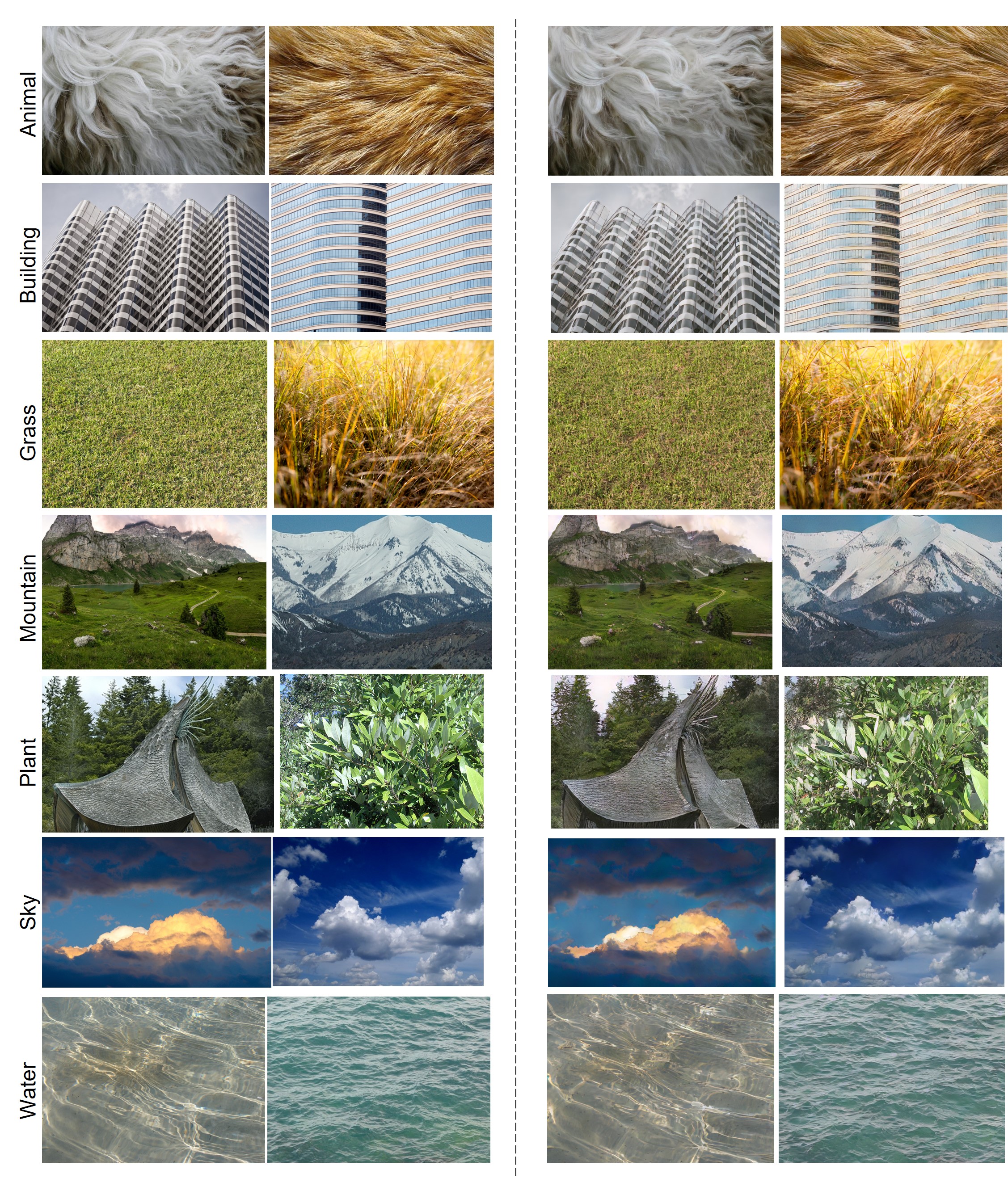}
    \caption{Reconstruction examples from OST dataset.}
    \label{fig:ost_rec}
\end{figure*}

\begin{figure*}
    \centering
    \includegraphics[width=.48\textwidth]{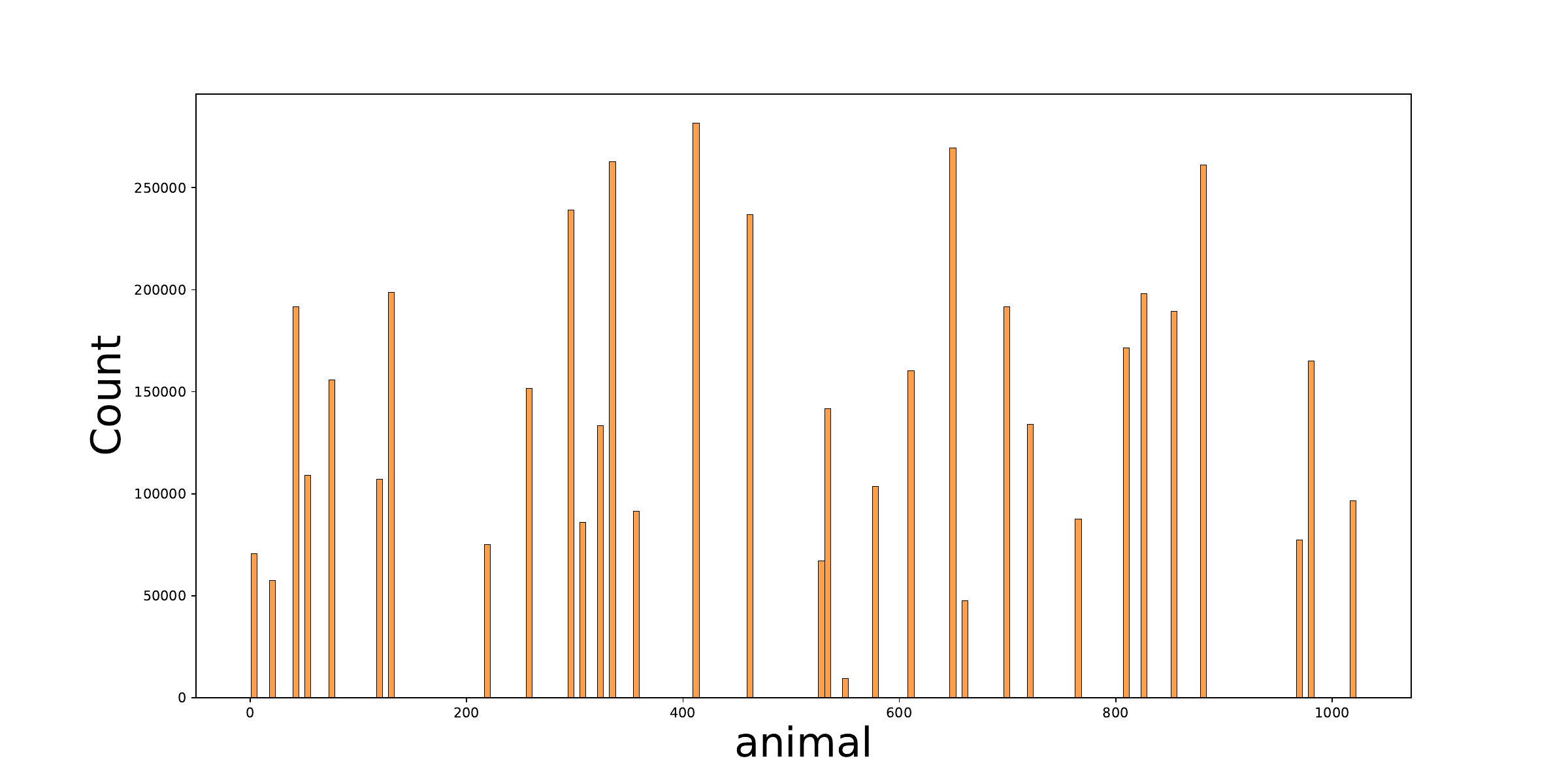}
    \includegraphics[width=.48\textwidth]{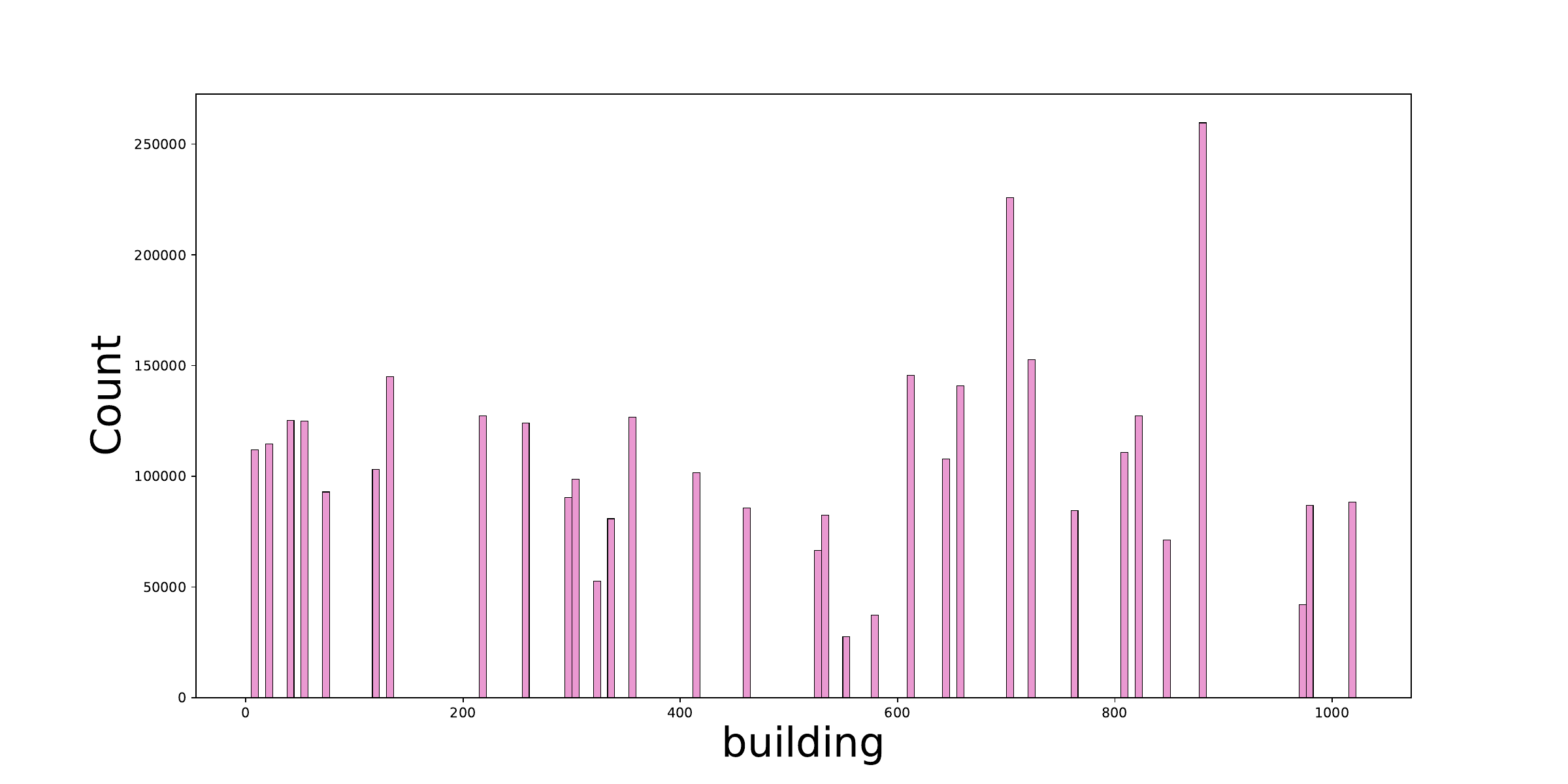} 
    \\
    \includegraphics[width=.48\textwidth]{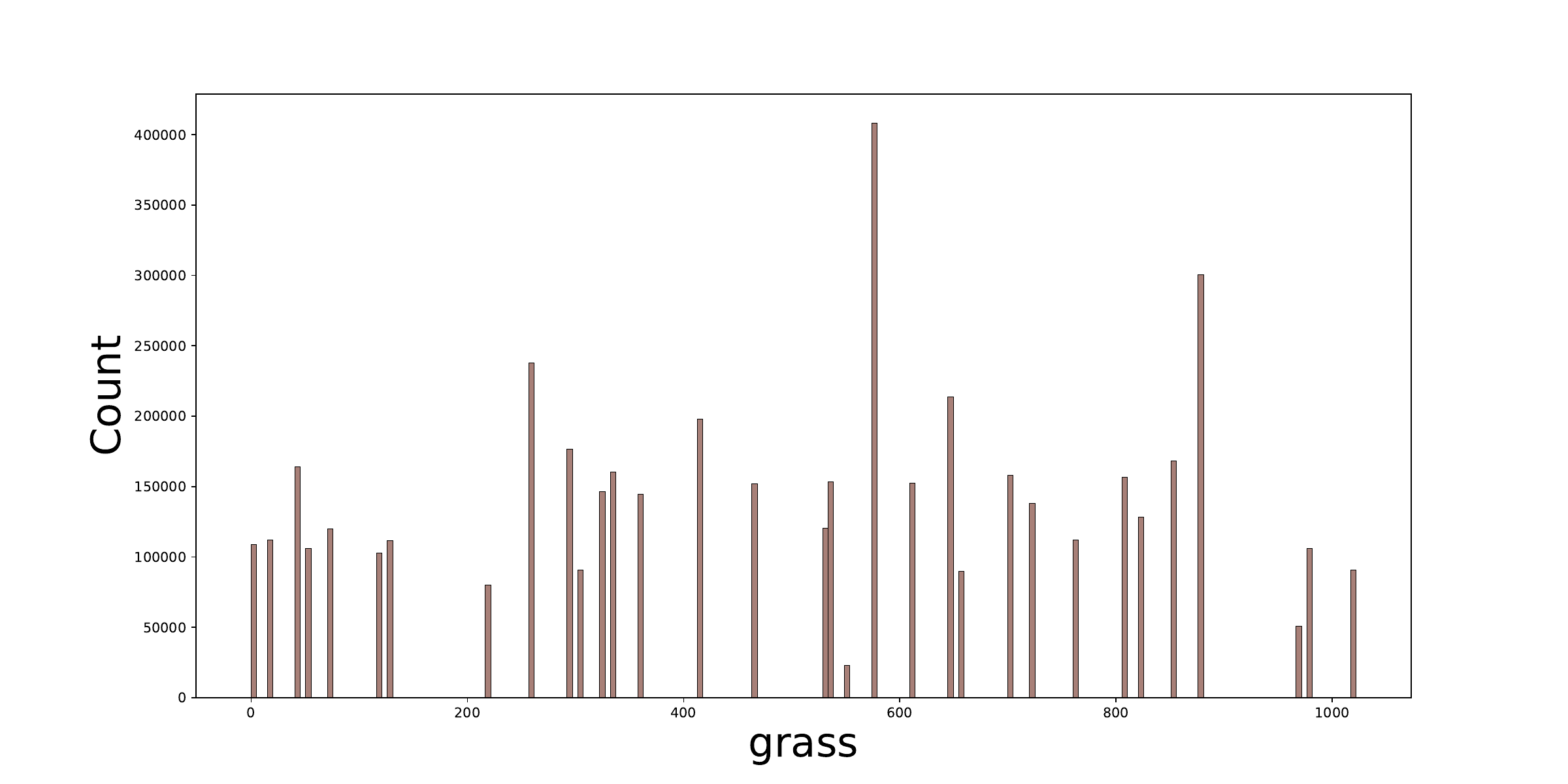}
    \includegraphics[width=.48\textwidth]{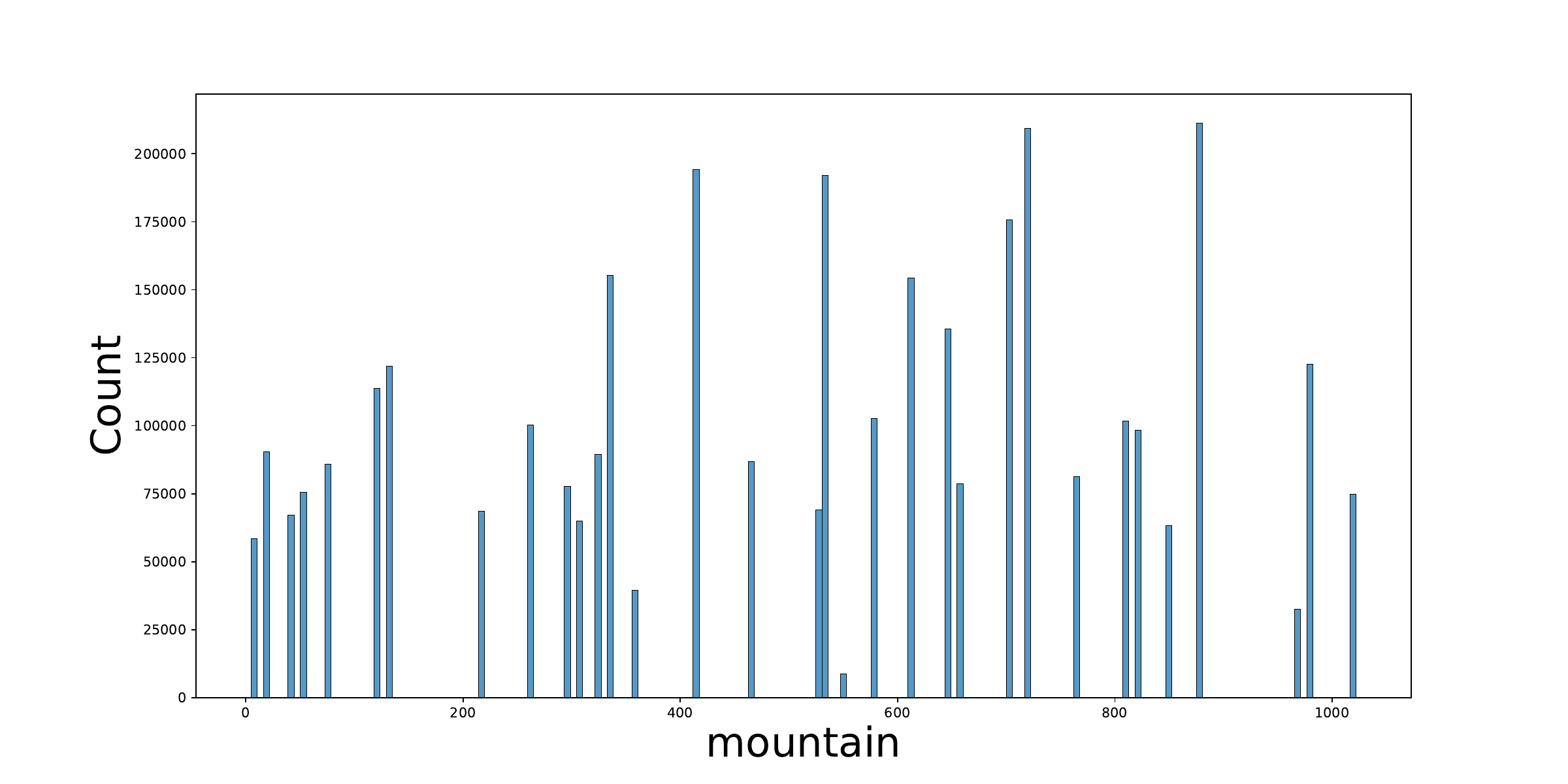} 
    \\
    \includegraphics[width=.48\textwidth]{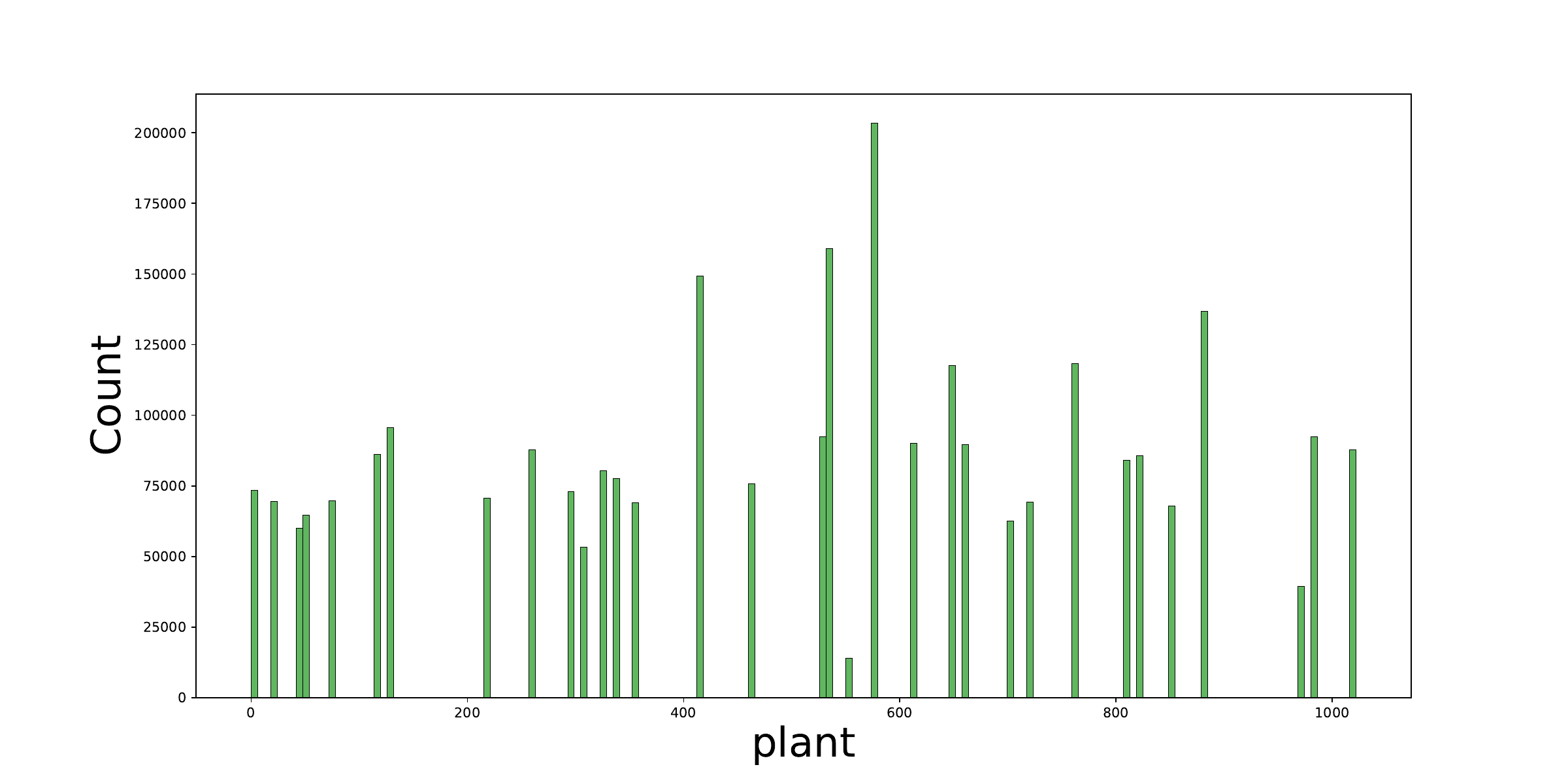}
    \includegraphics[width=.48\textwidth]{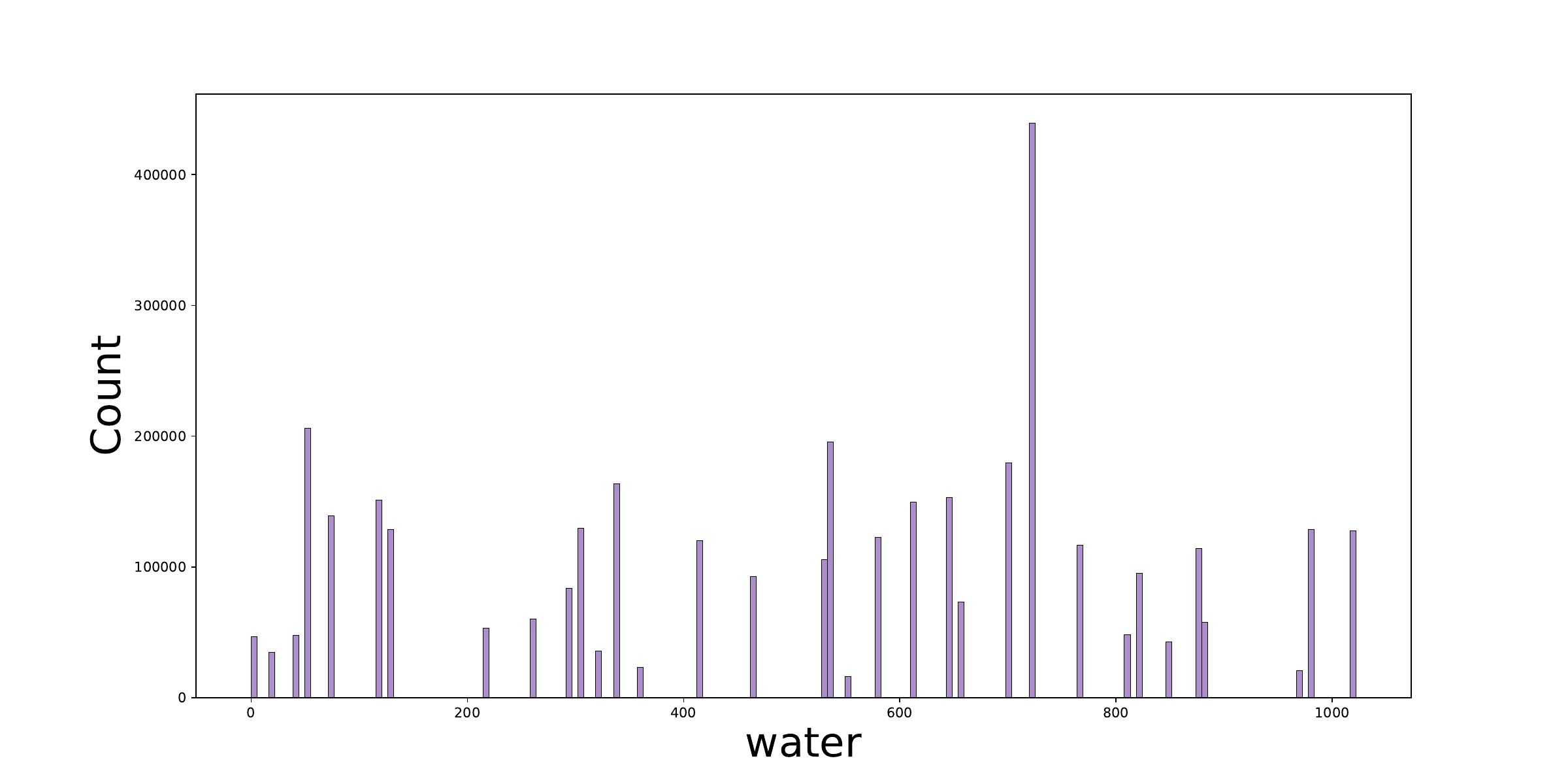} 
    \\
    \includegraphics[width=.48\textwidth]{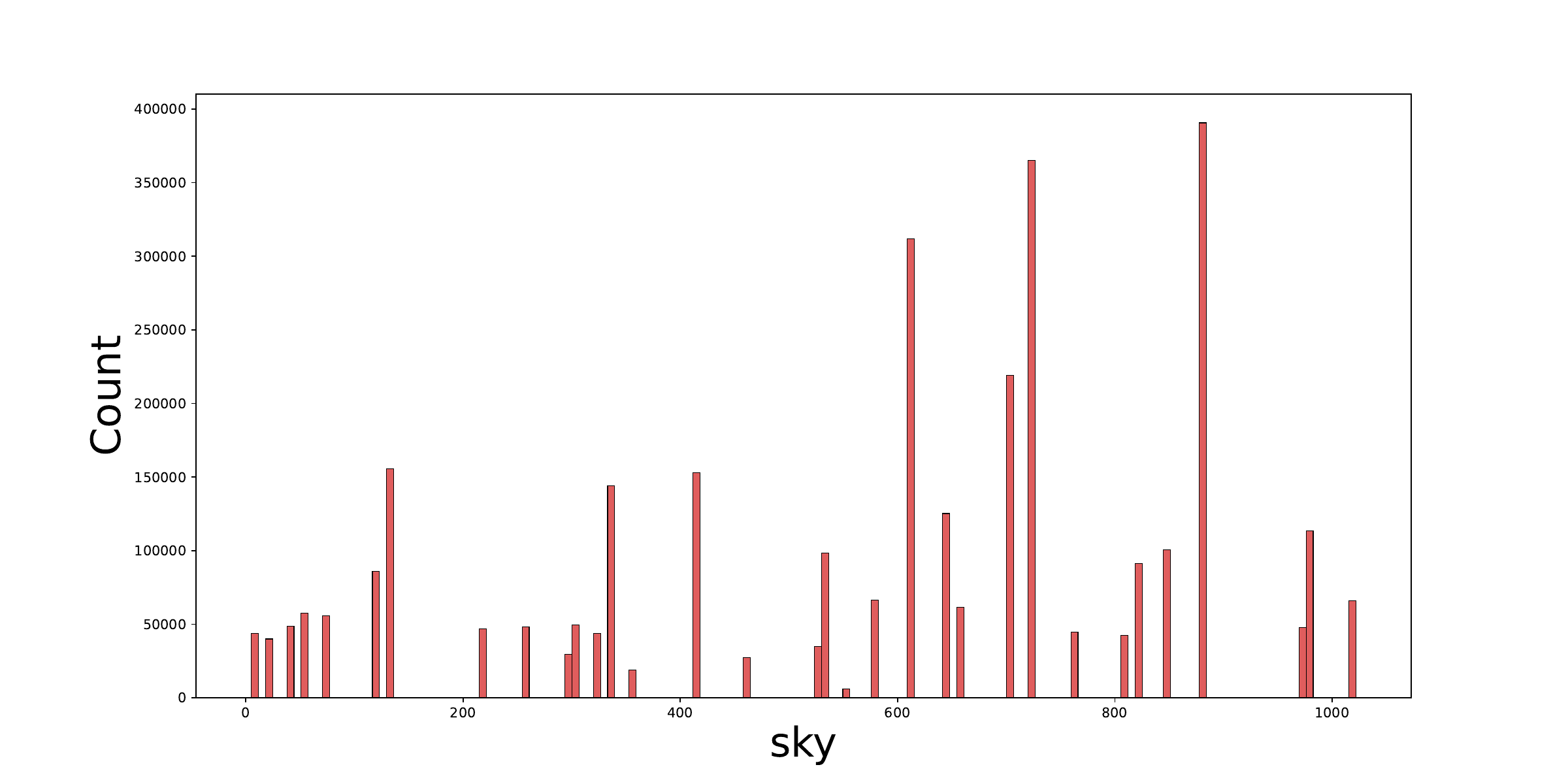} 
    \caption{Code index distribution for 7 texture categories in OST dataset.}
    \label{fig:code_dist}
\end{figure*}

\begin{figure*}[t]
    \centering
    (a) Animal
    \\
    \includegraphics[width=1.\textwidth]{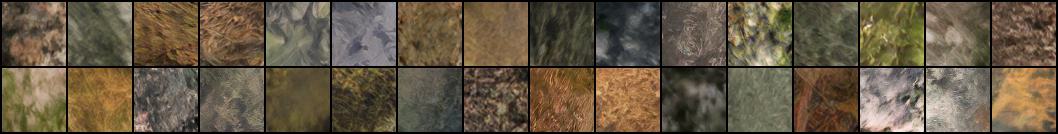}
    \\
    (b) Building 
    \\
    \includegraphics[width=1.\textwidth]{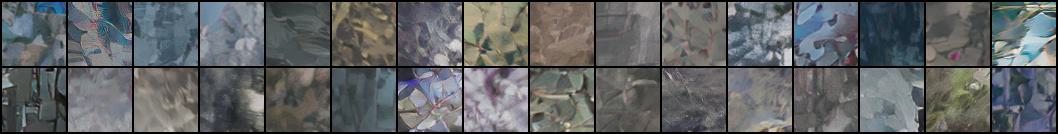}
    \\
    (c) Grass 
    \\
    \includegraphics[width=1.\textwidth]{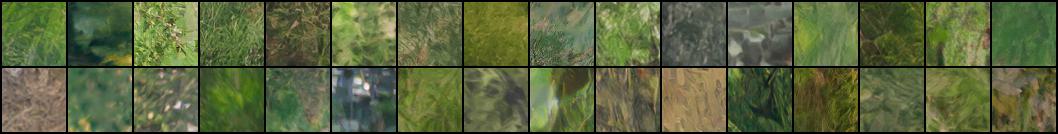}
    \\
    (d) Mountain 
    \\
    \includegraphics[width=1.\textwidth]{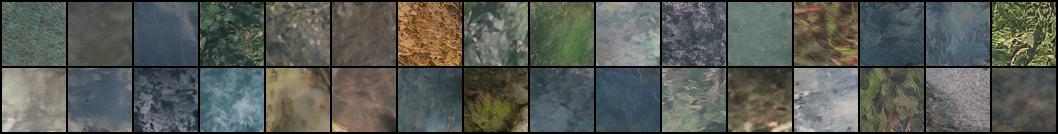}
    \\
    (e) Plant 
    \\
    \includegraphics[width=1.\textwidth]{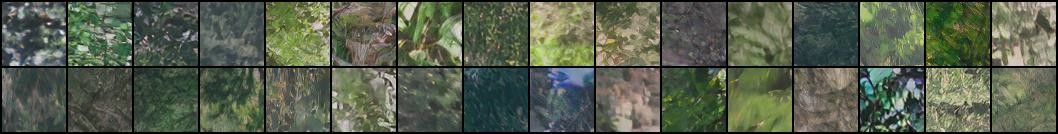}
    \\
    (f) Sky 
    \\
    \includegraphics[width=1.\textwidth]{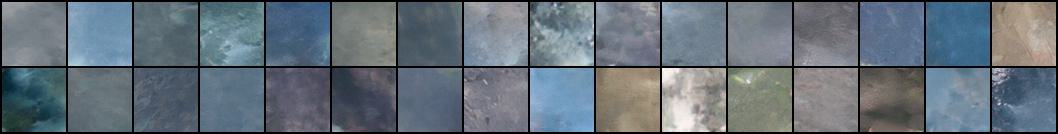}
    \\
    (g) Water 
    \\
    \includegraphics[width=1.\textwidth]{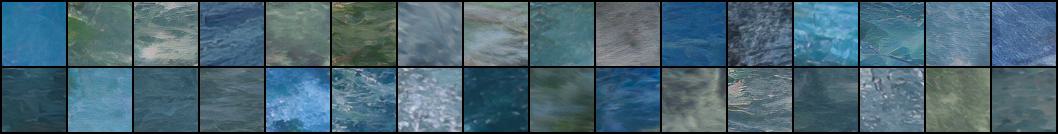}
    \caption{Visualization of different semantic-related textures encoded in HRP.}
    \label{fig:hrp_vis}
\end{figure*}

In this part, we show more empirical visualizations of the learned high-resolution priors (HRP). Firstly, we show an overview of all the 1024 codes separately in \cref{fig:codebook_overview}. Then, we show visualization of semantic-related codes with the help of a semantic texture datasets, \ie, the OST dataset \cite{wang2018sftgan} with the following steps:
\begin{itemize}
    \item Obtain the codes of each image with pretrained VQGAN. \Cref{fig:ost_rec} show some examples of the reconstruction results with our model.
    \item Calculate the distribution of each texture category on the codebook $\mathcal{Z}$, as shown in \cref{fig:code_dist}
    \item Sample codes from the distribution of each class and randomly arrange them to compose a $8\times 8$ latent feature $z$, and then decode it to $64 \times 64$ RGB texture patch. 
\end{itemize}
Finally, \Cref{fig:hrp_vis} empirically visualizes the learned HRP on OST dataset. We can observe that codes sampled from different texture distribution generates textures similar to the corresponding semantics, which proves the effectiveness of the HRP. Please note that the texture shapes are closely related with orders of latent codes, \cref{fig:hrp_vis} are just empirical statistic visualizations with random arrangement of codes. 

\subsection{Failure Cases}

By observing the results, we empirically discovered a limitation of our proposed FeMaSR: it favors natural textures over artificial textures, \eg the straight lines that dominate the building images.
Our method usually generates curved lines instead (see Fig. \ref{fig:limitations}). A similar phenomenon also occurs in neural texture synthesis \cite{gatys2015texture}. 
We leave the solution of this problem to future work.   

\subsection{Qualitative Results} \label{sec:vis_results}

\begin{figure*}[p]
    \centering
    \includegraphics[width=0.99\textwidth]{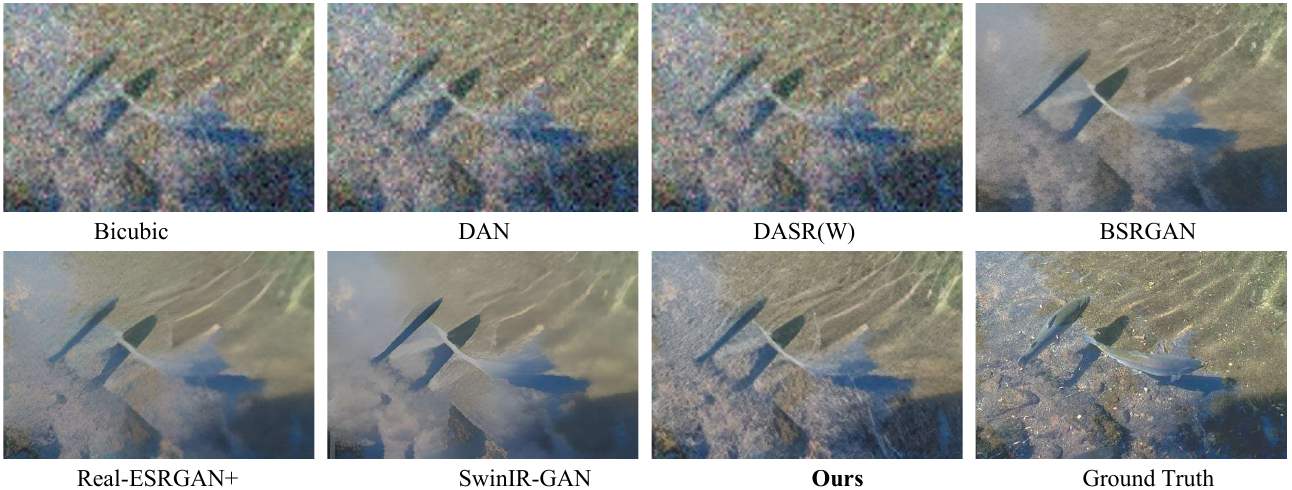}
    \includegraphics[width=0.99\textwidth]{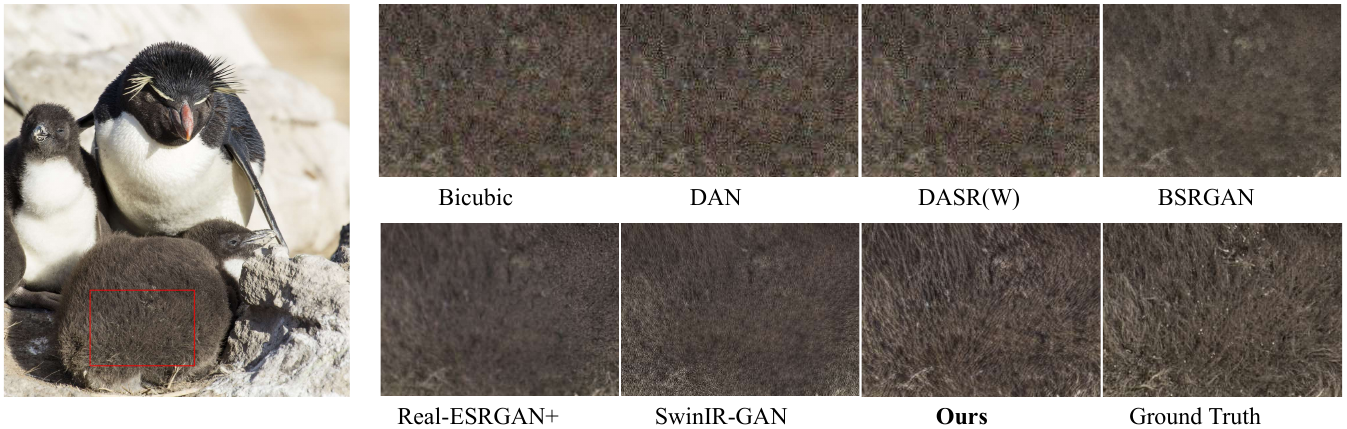}
    \includegraphics[width=0.99\textwidth]{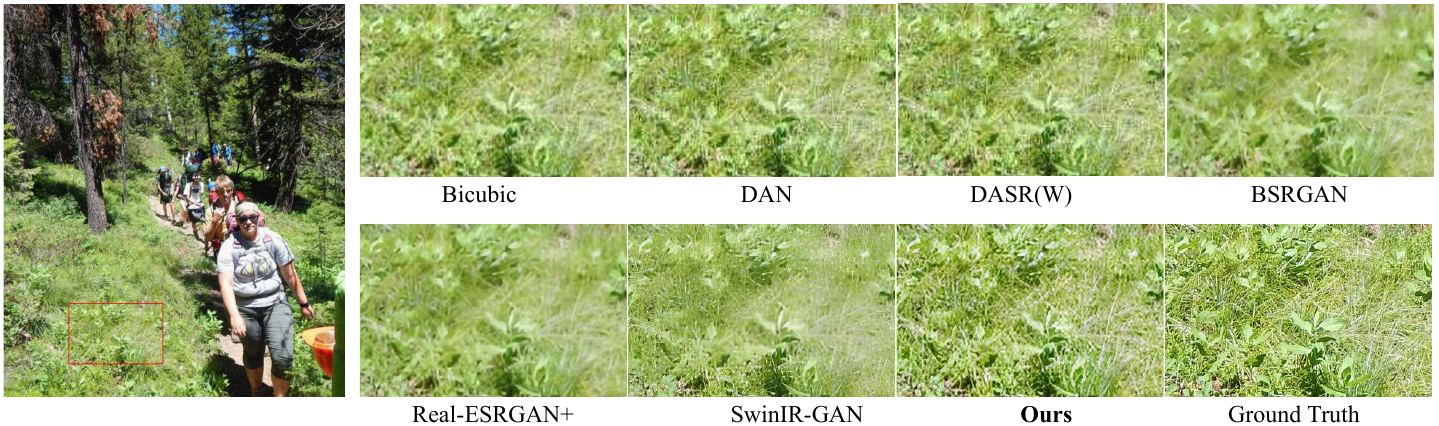}
    \includegraphics[width=0.99\textwidth]{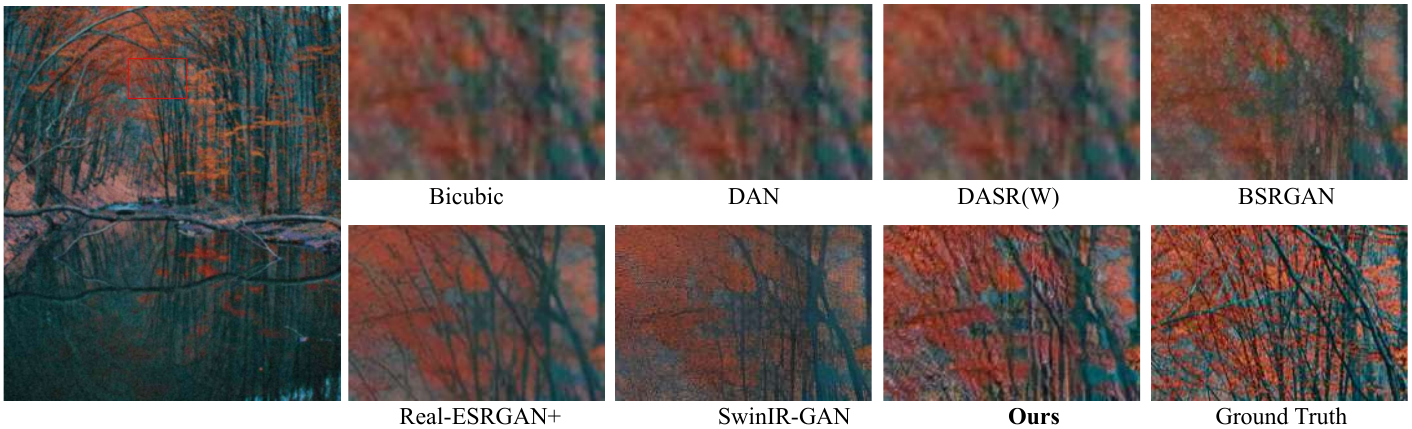}
    \caption{More results on synthetic benchmarks.}
    \label{fig:syn}
\end{figure*}

\begin{figure*}[p]
    \centering
    \includegraphics[width=\textwidth]{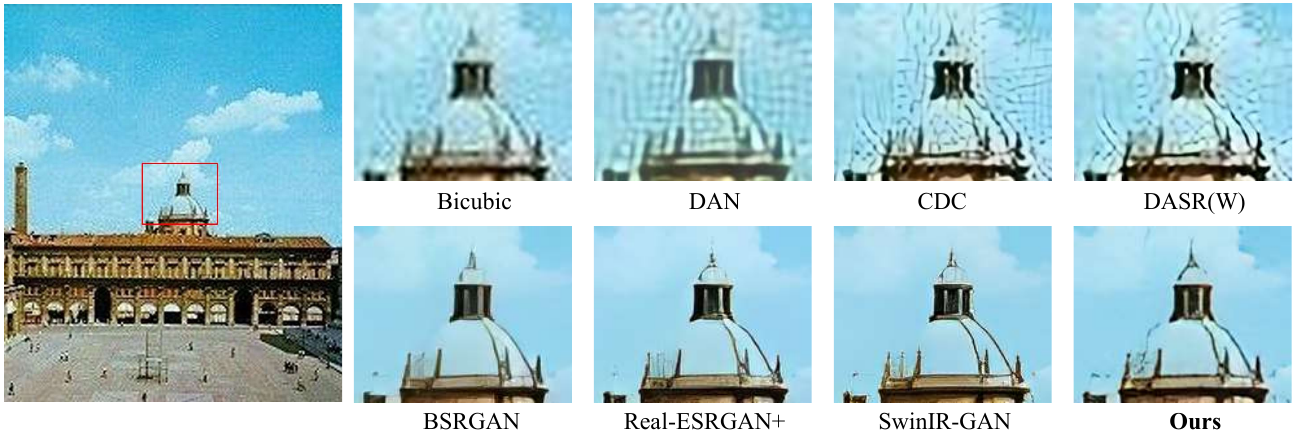}
    \includegraphics[width=\textwidth]{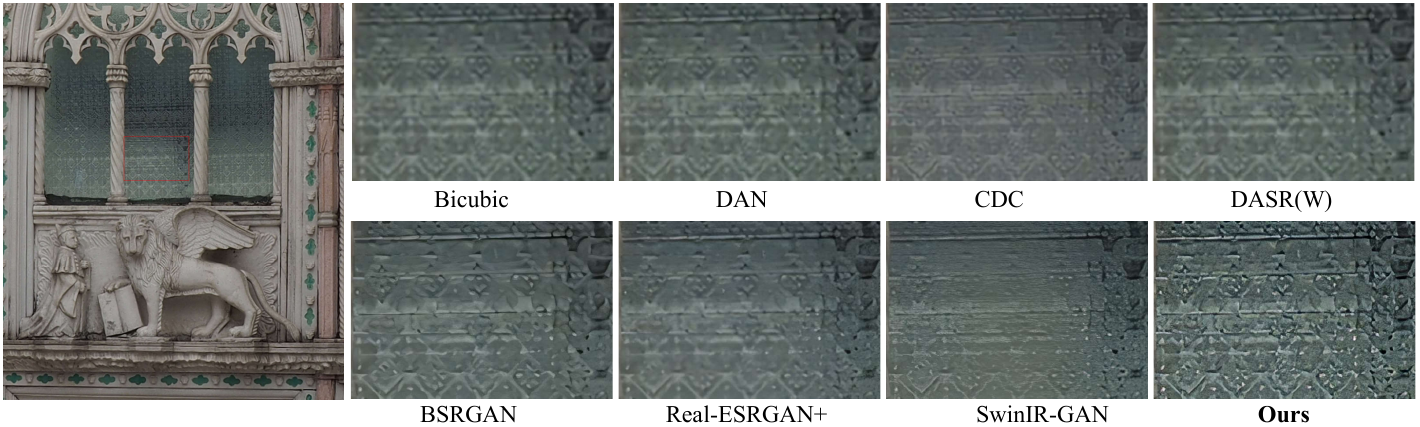}
    \includegraphics[width=\textwidth]{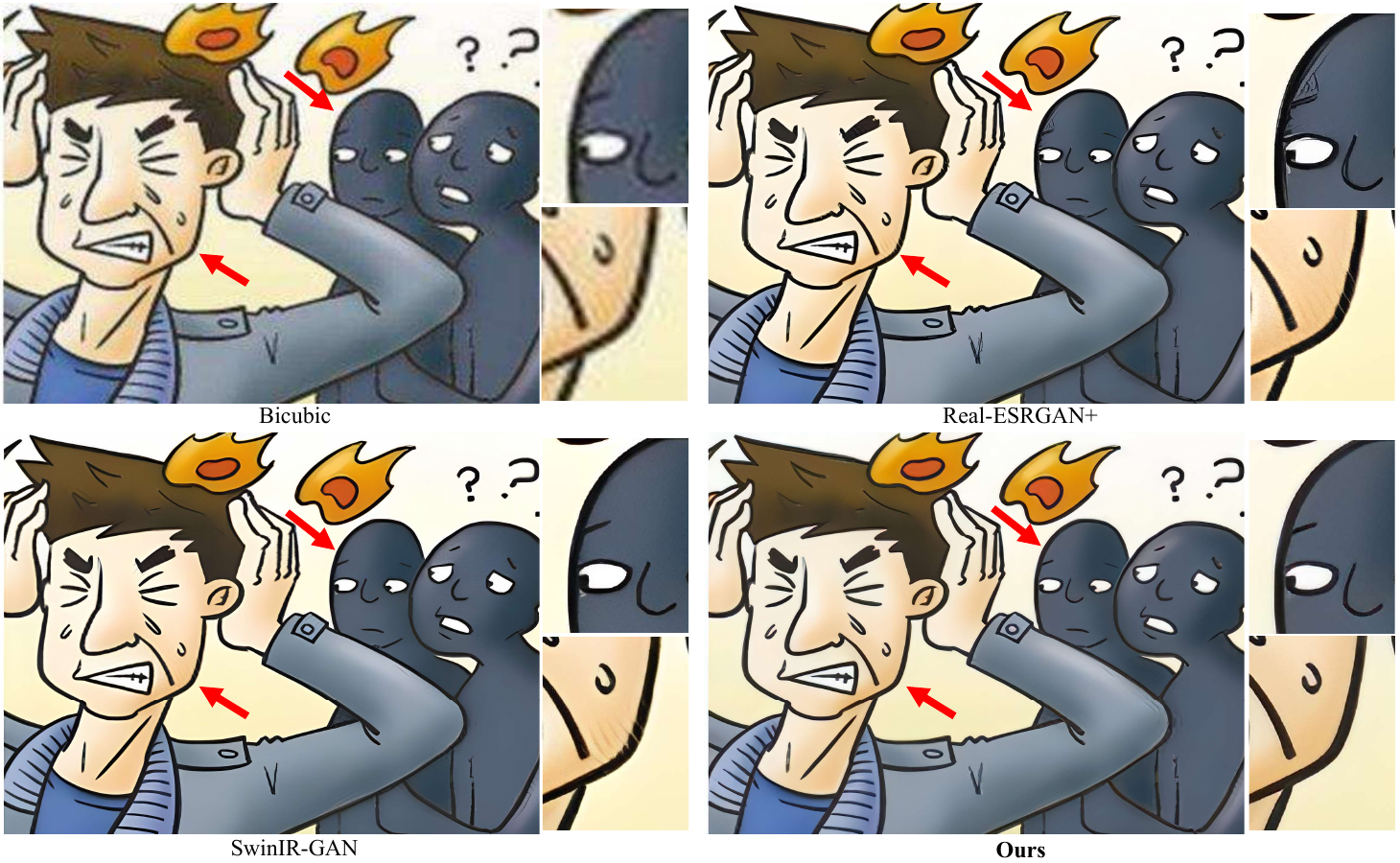}
    \caption{More results on real-world test images.}
    \label{fig:real}
\end{figure*}

We show more results on synthetic datasets in \cref{fig:syn} and real-world test images in \cref{fig:real} 

\end{document}